%% file: BIASinDE.tex
\documentclass[preprint]{elsarticle}

\InputIfFileExists{MISC/newCommandsAndPackages}{}{}

\begin{document}
\begin{frontmatter}
\title{Infeasibility and structural bias in Differential Evolution}

\author[uni1]{Fabio Caraffini\corref{corr}}\ead{fabio.caraffini@dmu.ac.uk}\ead[url]{www.cse.dmu.ac.uk/~fcaraf00}

\author[uni2,school]{Anna V. Kononova}\ead{anna.kononova@gmail.com}
\author[uni2]{David Corne}\ead{d.w.corne@hw.ac.uk}

\address[uni1]{Institute of Artificial Intelligence, School of Computer Science and Informatics, De Montfort University, Leicester, LE1 9BH, UK}
\address[uni2]{Department of Computer Science, Heriot-Watt University, Edinburgh, EH14 4AS, UK}
\address[school]{School of Communication, Media and Information Technology, Hogeschool Rotterdam, The Netherlands}

\cortext[corr]{Corresponding author}
\journal{Information Sciences}

\begin{abstract} 
This paper thoroughly investigates a range of popular DE configurations to identify components responsible for the emergence of structural bias -- recently identified tendency of the algorithm to prefer some regions of the search space for reasons directly unrelated to the objective function values. Such tendency was already studied in GA and PSO where a connection was established between the strength of structural bias and population sizes and potential weaknesses of these algorithms was highlighted. For DE, this study goes further and extends the range of aspects that can contribute to presence of structural bias by including algorithmic component which is usually overlooked -- constraint handling technique. A wide range of DE configurations were subjected to the protocol for testing for bias. Results suggest that triggering mechanism for the bias in DE differs to the one previously found for GA and PSO -- no clear dependency on population size exists. Setting of DE parameters is based on a separate study which on its own leads to interesting directions of new research. Overall, DE turned out to be robust against structural bias -- only DE/current-to-best/1/bin is clearly biased but this effect is mitigated by the use of penalty constraint handling technique. 

\end{abstract}

\end{frontmatter}

\section{Introduction}
Modern world is overwhelmingly confronted with optimisation problems of varying difficulty. The task of solving the majority of them exactly is unattainable due to particular features of an optimisation problem, such as extremely high dimensionality, absence of the analytical objective function, presence of constraints, computationally expensive evaluation of the objective function, etc. Thus, it is stochastic approaches for general purpose optimisation that come into play to find the "next best thing" -- the near-optimal solutions.
Amongst the available options, a common and logical choice of tools for such problems is from the field of Evolutionary Computation (EC) and Swarm Intelligence (SI). Most successful EC and SI algorithms happen to be population-based metaheuristics where the principle of natural evolution is taken as a source of inspiration to design the operators that select, mix and manipulate individuals that make up the population. The whole fields of EC and SI operate under the assumption that by representing candidate solutions of an optimisation problem in the form of these individuals, it is possible to progressively improve their quality through a simulated ``evolutionary'' process which eventually leads to identifying the near-optimal solutions. The most used frameworks that found their place in several engineering applications \cite{DEeng,bib:Miettinen99,bib:quagliarella97,PSOelectricaleng}, in system control and other industrial processes \cite{refId0,bib:LegoRobot}, in Robotics \cite{bib:spacerobot,robotsoccerEvolalg}, and in many other subjects and real world scenarios, are the Genetic Algorithm (GA)
, the Evolution Strategy (ES)
, the Particle Swarm Optimisation (PSO)
 and the Differential Evolution (DE) 
. 

However, since metaheuristics are not flawless, undesired algorithmic behaviours common for the majority of population-based algorithms can surface during the optimisation process. To illustrate this statement, premature convergence can prevent most evolutionary algorithms to explore the search space to an adequate extent. Such behaviour is quite common in GAs \cite{andre2001,leung1997degree} as a consequence of employing too exploitative adaptive operators \cite{bib:Rudolph2001} or due to the selection mechanism not being able to keep an adequate level of diversity in the evolving population. Similarly, stagnation can interrupt the exploratory process even in the presence of high diversity, as frequently occurs in DEs \cite{bib:Lampinen2000}. In fact, convergence is not generally guaranteed for the majority of the aforementioned stochastic algorithms: only a few have a general proof of convergence, see e.g. \cite{bib:Powel1964,Spall92multivariatestochastic,dorigo2002}, meanwhile in the vast majority of other cases proofs exist only under rather restrictive hypotheses \cite{algissue, Grippo2015ACO}.

Researchers are yet to find good explanations for these undesired behaviours, thus being unable to help general practitioners choose the most appropriate optimisation method. In other words, ``off-the-shelf'' solvers for specific classes of real-world problem do not currently exist. 
Therefore, challenges a practitioner encounters when faced with a real-world optimisation problem are not necessarily limited to those related to the nature or complexity of the optimisation problem itself. Often, lack of the theoretical knowledge on the dynamics inside general purpose algorithms, together with the absence of empirical indications on how to select and tune the most appropriate algorithms available in the literature, make it very difficult for the practitioners to be effective at solving general optimisation problems and for the computer scientists to be efficient in designing novel methods. 

Currently, a list of factors preventing the design of the ``ready-to-use'' algorithms varies a lot. From a \textit{technical} point of view, it is worth making the following considerations:
\begin{itemize}
    \item the No Free Lunch Theorem (NFLT) by Wolpert and Macready \cite{bib:NFL} states that a universal algorithms for the black-box optimisation cannot exist, and that problem specific information needs to be available to tailor algorithms to the problem; 
    \item even though the algorithmic design can be taken to a higher level by means of the hyperheursitc \cite{Burke2019} and memetic computing (MC) \cite{ong10} (paradigms, for which NFLT seems not to hold \cite{polilunches}), it is still a partially blind process due to the lack of theoretical knowledge regarding the internal dynamics of the nature inspired algorithms;
    \item the exploration/exploitation conundrum is still unsolved due to the necessity of performing the optimisation process under a limited computational budget in the real-world applications.
\end{itemize}
Moreover, from a \textit{practical} point of view it must be mentioned that: 
\begin{itemize}
\item the current tendency to iteratively improve upon previous algorithms, e.g. by employing multiple search strategies and self-adaptive operators as in \cite{bib:Brest2006a,cheng2015}, has led to the emergence of over-complicated optimisation frameworks carrying a high algorithmic overhead and thus, being less suitable for implementation in devices with limited memory or less inadequate for real-time problems and large-scale tasks \cite{bib:LegoRobot}; meanwhile also being more difficult to tune; 
\item hybrid approaches, from the field of Memetic Computing or hyperheuristics in particular, are full of examples of heavy algorithmic skeletons as in \cite{bib:deOca2009,bib:Molina2010,bib:Peng2010}, where multiple algorithms are misguidedly merged and therefore requiring ``meta-optimisation'' over a high number of artificial parameters \cite{Nannen:2006, MASON2018}.
\end{itemize}

To overcome these issues, researchers have recently started investigating possible ways for tuning parameters, either off-line \cite{EIBEN201119} or on-the-fly 
\cite{bib:Brest2006a}, to save the computational budget and processing time. This is usually done via optimising surrogates of the actual objective functions \cite{bib:OngMultSurrogate2007}, or more recently, by designing fitness landscape analysis methods to be executed prior to starting the optimisation process \cite{CaraffiniSPAM2014,malan2013survey,bib:poikocluster}. Nonetheless, a deep understanding of the dynamic inside the populations of the EC algorithms has not matured yet, and the need to understand how individuals process and propagate information inside the population of a nature inspired algorithms is still unsatisfied. Thus, notwithstanding the fact that a lot of research has gone into EC methods, the causes of the aforementioned undesired algorithmic behaviours are still unclear and need to be clarified based on further studies of the interactions amongst candidate solutions inside the evolving population and the corresponding search in the decision space. 

To shed light on these open research questions, recent studies have focused their attention on the variability of individuals inside the populations and have found biases in popular metaheuristics such as GA and PSO \cite{KONONOVA2015}. The presence of such bias in the search process has been established to be intrinsic to the algorithmic structure and the search logic, as well as correlated with the common parameters, first and foremost -- the population size. Implications of such discovery are multiple and positively impact the algorithmic design process. Lately, a preliminary study has extended what has previously been done in \cite{KONONOVA2015} to some DEs \cite{bib:biasDELego18}. This study has unveiled interesting links between the structural bias and particular combinations of the DE operators that are worth further investigation and extension to all popular DE variants. The latter is therefore done in this paper.

The remainder of the paper is structured as follows:
\begin{itemize}
    \item Section \ref{background} summarises the literature on algorithmic bias, presents previous results and test-beds to look for biases inside the population-based algorithm as well as discusses how correction strategies, to handle solution generated of the search space, could contribute to this phenomenon by adding their own bias when embedded in some algorithmic contexts;
    \item Section \ref{DE} briefly presents the Differential Evolution framework and $4$ of its most popular scheme used in this piece of research;
    \item Section \ref{OandM} clarifies the objectives of this study and gives details on the methodology and experimental setup required to address them and reproduce the presented results;
    \item Section \ref{results} comments on the results;
    \item Section \ref{conclusions} concludes the study outlining its general message and highlighting strengths and weaknesses of the proposed approach.
\end{itemize}

\section{Background}\label{background}
An algorithm is said to \textit{possess structural bias} when its individual parts put altogether are unable to explore all areas of the search space to the equal extent, irrespectively of the fitness function \cite{KONONOVA2015}. When faced with the task of optimising a given function, a general-purpose optimisation algorithm is expected to be able to locate the optima regardless of where they are located in the search space. Following the general idea of iterative optimisation algorithms, values of the objective function in the points that are the current candidate solutions pull the search in the direction of improvement as prescribed by the algorithm at hand. However, as has been established in \cite{KONONOVA2015}, combination of individual operators of the algorithm also pulls the search in the direction of their individual biases. Thus, optimisation process ends up being a complex \textit{superposition} of two not-necessarily-agreeing pulls: the unknown "evolutionary" pull originating from the incremental re-sampling of candidate solutions and their "fitnesses" and the unknown pull stemming from the ''sum'' of individual biases of algorithm's operators when applied to the current candidate solutions. Theoretically, in this superposition of pulls, the first one is dominated by the objective function, meanwhile the second is predominantly defined by the algorithm's structural bias.

For an algorithm not to exhibit structural bias, its generating operators should be able to reach every part of the search space without imposing any preferences on some regions of the domain over others. If this behaviour is not sustained, the search process becomes biased and a thorough analysis should be carried out to identify the triggering mechanisms for the manifested structural bias. Clearly, different combinations of functions, domains, constraints and corresponding constraint handling strategies impact the strength of the bias differently and, thus, greatly complicate its study by introducing multiple highly interlaced aspects to be considered simultaneously. Therefore, an adequate protocol must be defined to be able to search for algorithmic biases. After commenting previous results on bias in Section \ref{previousResults}, the methodology previously established to produce the results presented in this article is described in Section \ref{testingForBias}. 

\subsection{Existing results on bias}\label{previousResults}
A first attempt at pointing out the presence of structural biases has been made by the PSO community. The empirical observation that some PSO algorithms were incapable of exploring too far from the origin of the search space, i.e. so called ``centre-seeking bias'' (CSB), and also too far from the initial swarm of candidate solutions, i.e. ``initialisation-region bias'' (IRB), have attracted attention of researchers and paved the way for more thorough analyses. Initially, a great deal of attention has been paid to finding the mechanism to avoid such limitations of the algorithms. The study in \cite{bib:Angeline1998} has introduced selection operators to relax the rigid ``one-to-one-spawning'' logic that at the time has been thought to be the source of such undesired phenomena. However, little effort has been made to understand and define its actual nature -- a proper formalisation for CSB and IRS has been finalised later on in \cite{DAVARYNEJAD}. In the meantime, controversial studies, as e.g. \cite{bib:Monson2005}, speculating the existence of CSB in all population based algorithm have started looking at different optimisation paradigms. The latter, confuted in \cite{bib:Kennedy2007}, has been followed by further investigations on PSO as e.g. 
\cite{psorot2007} where it has been showen that, unlike DE \cite{bib:rotationEvo,rot2018}, PSO seems to be sensitive to rotations of the objective functions. In this light, it is worth mentioning that an ``angular'' bias can also be defined for PSO \cite{Spears}.

Not long after the latest results in \cite{DAVARYNEJAD}, a generalised definition of ``structural bias'', applicable to all population-based metaheuristics, has been given in \cite{KONONOVA2015} where the PSO paradigm has been proven to be plagued by an intrinsic structural bias, with both theoretical and empirical evidence on such phenomenon. Furthermore, the study has also considered a simple GA thus adjusting the erroneous assumption made in \cite{bib:Monson2005} (i.e. all population-based algorithms have centre-seeking bias) and showing how also optimisation algorithms equipped with selection and genetic operators can display a biased search logic (in contrast with the method proposed in \cite{bib:Angeline1998} where selection was employed to ease the strength of biases, with no justification of such choice). Most importantly, this piece of research has shown, under some hypotheses, that such structural bias in GAs correlates with the population size and since the latter being the common denominator for most population-based algorithms for real-valued global optimisation, structural bias is amplified proportionally to the population size. Such conclusion was empirically validated, thus confronting the common belief that a large population size is beneficial for a number of reasons, e.g. a summary is given in \cite{bib:Bennett2010}, and is a key to tackling large-scale problems efficiently \cite{tuningLampinen}. Finally, a graphical approach to visualising structural bias, as a non-uniform clustering of the population over time, was developed. 

The visual approach for representing the structural bias in \cite{KONONOVA2015} has turned out to be extremely practical and largely used by researchers in the field who have recently adopted this methodology e.g. to individualise and then remove structural biases from the so called JADE and SHADE-based algorithms, see \cite{PIOTROWSKI2018}, and to show that the current state-of-the-art is mainly represented by the overcomplicated algorithms which should be simplified by removing their bias via a second, more informed, algorithmic design process \cite{Piotrowski2016,PIOTROWSKI201832}.

It must be pointed out that the latter articles focus on removing or studying some particular DE variants, but a comprehensive analysis on the structural bias in DE has only been started later in the preliminary study described in \cite{bib:biasDELego18}, which is extended here.

\subsection{Testing for bias}\label{testingForBias}
The most suitable choice of testbed for the identification of the structural bias, in terms of its effects on the distribution of the final best solutions over multiple runs, is a series of experiments, for chosen fixed dimensionality, on the function
\begin{equation}\label{eq:f0}
    f_0: [0,1]^n \subset \mathbb{R}^n \to [0,1]\subset\mathbb{R},\quad f_0(x)=\text{Uniform}\left(0,1\right)
\end{equation}
for which, as rigorously explained in \citep{KONONOVA2015}, an ideal unbiased algorithm should return a uniform distribution of the best final solutions, over a series of independent runs. 
In this sense, $f_0$ serves as a reference problem which \textit{by design allows a decoupling between artefacts of the objective function and artefacts arising from iterative application of algorithmic operators} as value of $f_0$ in any point does depends neither on the values within its neighbourhood, nor on the past evaluations of this point. This is achieved by effectively eliminating the influence of the local positions of the candidate points but retaining the underlying algorithmic artefacts.

It must be remarked that the discussion above holds for all population-based algorithms, regardless of the differences in their algorithmic components. This includes Differential Evolution where a parent selection is missing (it employs the ``one-to-one-spawning'' mechanism as does PSO, which seems to be biased), but individuals are necessarily selected at the mutation level either at random, or according to a stochastic rank-ordering over a specific set of values of the objective function in the current population. 

Procedure for testing an optimiser on $f_0$ consists in repeating optimisation over a pre-established fixed number of independent runs and studying the resulting distribution of the final best solutions per run. A value of $50$ runs has been used in our earlier studies to guarantee statistically significant results. A practical visual approach, originally proposed in \cite{KONONOVA2015} and successfully re-employed in \cite{bib:biasDELego18}, consists in displaying the obtained final distribution of solutions in ``parallel coordinates'' \cite{Inselberg1985,KONONOVA2015}: coordinates of a vector in an $n$-dimensional space are marked correspondingly on $n$ equally spaced parallel lines. Thus, an unbiased algorithm would yield a figure with points homogeneously filling the entire interval in each dimension. Conversely, in the presence of a strong structural bias, clusters will appear as best solutions would tend to accumulate in one or more segments of each parallel line.

\subsubsection{Random generator effects}
It is worth mentioning that an extensive study has been carried out in \cite{KONONOVA2015} to demonstrate that structural bias cannot be attributed to the artefacts of the random generator used. The essence of this study consist in assuming, by contradiction, that there exists a correlation between random numbers used to generate coordinates of the two subsequent points examined by the algorithm, and that correlation is significant enough to measure. To find such correlations between elements of the pseudorandom sequences, thus proving the impact of the random number generator on the structural bias, three tests were designed and executed. The first test was run to examine the correlation between consecutive pairs of random values used to generate some specific dimension values of points in the search space. Similarly, the second test was run to examine the correlation between all dimensions simultaneously. Finally, the third  test was run to track the correlation between consecutive values (as in the first test) in the whole pseudorandom string. Full details are available in \cite{KONONOVA2015}.

No evidence of such correlation was found by neither of the tests and conclusions of the study unequivocally suggested that observations of structural bias by means of evaluation on $f_0$ do not originate from the random generator but rather represent artefacts from the iterative application of the algorithmic operators.

To conclude, it has to be remarked that all algorithms discussed in the current publication employ the same Java $48-$bit pseudorandom generator as in \cite{KONONOVA2015}, which is based on the linear congruential generator (LCG)  \cite{bib:LEcuyer1999} with a period of $2^{48} \approx 2.8\times10^{14}$ and the seed automatically generated by means of the system time routine \textit{System.currentTimeMillis()}. Thus, the authors find unnecessary to redo such analysis for the current paper -- this question is considered settled unless a different random generator is used.

\subsection{Constrained problems}\label{constrained}
 To complicate things further, a great deal of optimisation problem is explicitly or implicitly constrained: ranging from simple inequalities that produce a hypercube domain through to the complex disconnected sets defined through simulations or as solutions of large systems of complex equations. Most constraint optimisation problems can be considered as ambiguously defined since function values outside the domain usually tend not to be specified. Unfortunately, constraint handling is not straight-forward in EAs as traditional variation operators are blind to constraints \cite{bib:EibenSmith}. This means that feasibility of parents does not guarantee feasibility of solutions they generate -- numerical variation operators are generally completely unaware of the boundaries of the search domain. Therefore, unless a very specific restrictive operator or encoder which somehow exploits regularities of the feasible search space  \citep{bib:Michalewicz1995} is used inside the algorithm to ensure the feasibility of the solutions inside the population, a strategy must be chosen on how to deal with solutions generated outside the domain at any stage of the algorithm. Thus, reduction of the amount of time spent generating infeasible solutions becomes an additional (possibly, implicit) \textit{objective for the algorithmic design}. 
 
 Over the years, a variety of strategies has been developed -- a summary of general state-of-the-art constraint handling techniques is reported below \citep{bib:Michalewicz1995}:
 \begin{itemize}
    \item \textbf{Penalise}: fitness value of a newly generated solution outside of domain is substituted by a predefined penalty value (binary, distance-based, time-dependent or some other adaptively calculated value); 
     \item \textbf{Dismiss} (death penalty): if a newly generated solution is outside the domain, it is dismissed and either re-generated or replaced by one of the parent solutions; 
     \item \textbf{Correct} (repair): based on a solution outside the domain, generate a new feasible solution according to the chosen method (saturation, toroidal, local search, etc.);
 \end{itemize}
Probabilistic formulations where constraint handling techniques are used over a proportion of solutions or modified solutions enter the population with some probability have also been investigated but have largely fallen out of fashion \citep{bib:Michalewicz1995,bibOrvosh1993}. Advantages and disadvantages of general use of the aforementioned techniques are summarised in Table \ref{tab:corr_strategies}. It must be remarked that constraint handling operators, undeniably being part of the algorithmic design, can contribute towards the algorithmic bias. However, current level of understanding of formation of bias is not sufficient enough to pinpoint their impact. 

\InputIfFileExists{TABS/corrStrat}{}{}
Furthermore, rather generally, all \textit{correction strategies} can be further classified as either superficial or complete ones. \textit{Superficial correction strategies} limit their impact to only reassigning coordinates of the out-of-domain points to be back within the domain without re-evaluating the resulting corrected point. Solutions corrected with such become in some sense ``alien'' to the original objective function -- they are the artefacts of the design choice (a choice of correction strategy) rather then the objective function itself -- and comparison of two algorithms employing drastically different correction strategies is akin comparing apples and the so-called ``Chinese apples''.

\begin{figure}[H] \centering
\subfigure[Solution out of domain ]{\includegraphics[width=.28\linewidth]{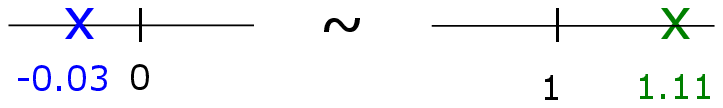}}
\subfigure[saturation correction result]{\includegraphics[width=.34\linewidth,trim={15mm 263mm 120mm 5mm},clip]{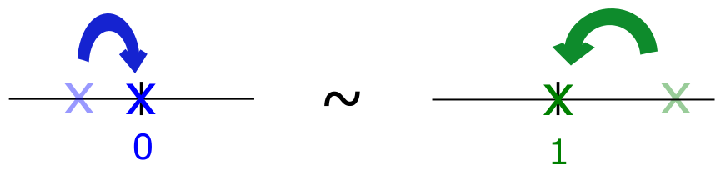}\label{fig:sketch_sat}}
\subfigure[toroidal correction result]{\includegraphics[width=.36\linewidth,trim={15mm 263mm 120mm 5mm},clip]{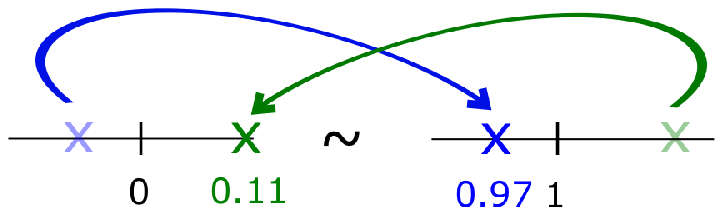}\label{fig:sketch_tor}}
\caption{Schematic explanation of correction strategies for domain $[0,1]$}\label{fig:correctionsGraphical}
\end{figure}

Schematic explanation of examples of superficial correction strategies popular in the field of EA are shown in Figure \ref{fig:correctionsGraphical}. 

Meanwhile, the \textit{complete correction strategies} actively search for other feasible inside-the-domain points in the (relative) vicinity of the currently unfeasible point. This search is typically performed via some local search routine with prescribed complexity and budget of function evaluations. Thus, complete strategies ensure that corrected solutions fully represent the original objective function unlike the superficial strategies that only assign artificially fitness values to moved-previously-unfeasible points. However, complete correction strategies create ``holes'' in the domains that are filled with unfeasible points for which no quality information is provided to the optimisation method. This directly contradicts a long-standing idea of the EAs operating via combining partial information from \textit{all} the population \cite{bib:Michalewicz1995}. Use of complete correction strategies introduces further problems such as being extremely time-consuming or a possibility of transferring limitations of the chosen local search algorithm into a failure to find suitable solutions by the overall method. Moreover, there is a risk that domain boundaries end up being over-explored due to the locality of the search concentrated around the boundary region\footnote{This highlights a subtle connection between the choice of correction strategy and structural bias. Results presented in this paper lead one to believe that, potentially, correction strategy plays an important role in the formation of structural bias of the method}. In practice, mostly due to time limitations and unwillingness to introduce additional design choices, complete correction strategies are rarely used. Instead, \textit{under a widely spread assumption of the non-importance of the correction strategy}, simple superficial corrections are typically preferred with no justification. We argue that such assumption is, in fact, \textit{erroneous}. 

We argue that in practice \textit{significantly more solutions} end up requiring correction than what is usually assumed by an algorithm designer. The overall number of corrections required during the optimisation provides a (rough) estimate for a degree to which the actual function being optimised corresponds to the original function, thus, projecting the justification to use local information. In other words, \textit{an overly high number of corrections deprives the algorithm from using information about the local structure of the landscape under investigation in most critical boundary regions}. 

\begin{figure*}[bt] \centering
\subfigure{\includegraphics[width=.72\linewidth,angle=270,trim={3mm 9mm 0mm 1mm},clip]{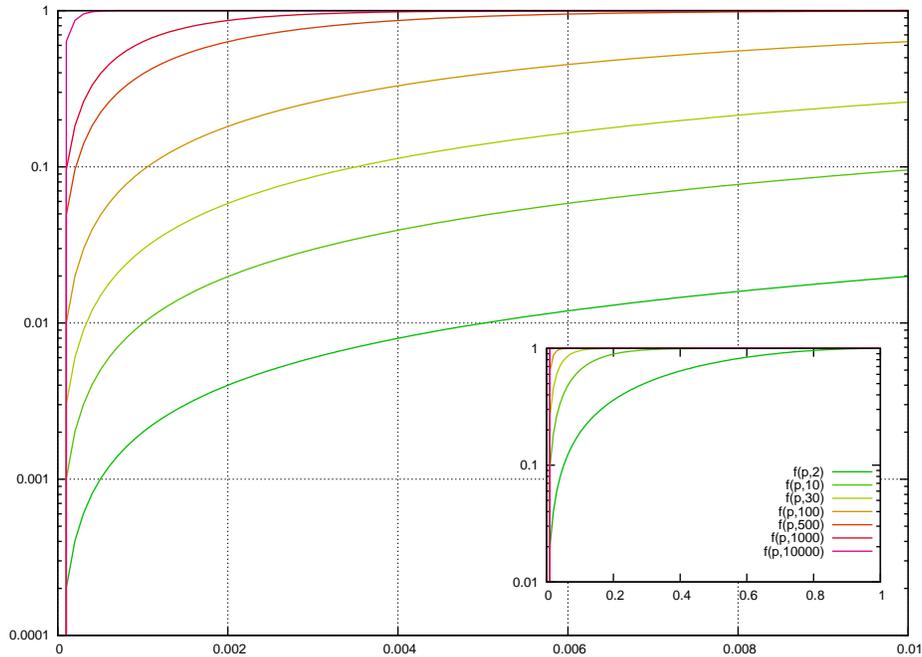}}
\caption{Tabulation of the probability of a solution requiring a correction in at least one dimension, as expressed by $f(p,n)=1-(1-p)^n$. Multiple curves correspond to different values of dimensionality $n$ defined for values of the constant rate $p$ at which corrections become necessary, independently for different dimensions (shown in horizontal axis). Note how fast probability gets close to $1$ for values of dimensionality over $30$. Zoom out to $[0,1]$ is shown in smaller figure in the same colours.}\label{fig:p0tab}
\end{figure*}

Clearly, a number of corrected solutions in the population grows with dimensionality of the problem since a solution with only one coordinate outside the domain already requires a correction. For a simplified experiment, suppose jumps outside the domain in different dimensions occur independently and with a constant rate $p\in[0,1]$\footnote{Obviously, this is a simplification of the real situation as this rate is not constant and there might be dependencies between dimensions.}. Then probability of a solutions to require corrections in at least one dimension is then trivially expressed as $f(p,n)=1-(1-p)^{n}$, where $n$ is problem's dimensionality. To help with visualisations, Figure \ref{fig:p0tab} shows tabulations of this formula for various values of dimensionality (values of $p$ are shown in horizontal axis and $f(p,n)$ -- in vertical). It is easy to see that $f(p,n)$ attains high values for relatively low values of $p$ and this effect is drastically increased as dimensionality $n$ increases. For a rather modest (by modern standards) value of dimensionality of $n=100$, the probability that at some iteration a correction is required in at least one dimension exceeds $0.99$ for $p>0.045$. As dimensionality increases further, virtually all solutions end up requiring corrections for majority of reasonable values of $p$. This clearly demonstrates the importance of the correction strategy choice for high dimensional problems -- \textit{thoughtlessly chosen correction strategy can transform the original optimisation problem into a trivial one which structurally has little to do with the original function}.

Depending on the nature of the problem, some constraint handling technique choices might even become prohibitive: e.g. impossibility of keeping unfeasible solutions in the population in case of domain boundaries from physical limitations of some equipment used to compute objective function values. Thanks to design, some strategies may lead to identical behaviour for some algorithms (e.g. penalty and dismiss in DE, as explained in Section \ref{OandM}). For other strategies, performance of an algorithm can turn out to be drastically different with different constraint handling strategies as clearly demonstrated by results presented later in this paper. However, despite trivially achievable conclusions above, usually, the choice of constraint handling technique is not investigated thoroughly during the design of an optimisation algorithm. We aim to draw the attention of algorithm designers and practitioners to the importance of this choice. \textit{It turns out that as usual, it matters what happens on domain boundaries}.

\section{Differential Evolution}\label{DE}
Differential Evolution is a powerful yet simple metaheuristic for global real-valued optimisation which only requires three parameters to function efficiently \cite{bib:Storn1995}: the scale factor $F\in[0,2]$, the crossover ratio $Cr\in[0,1]$ and the population size $NP$. The DE framework consists of the iteration of only three steps, as shown in Algorithms \ref{alg:DE}. In the sense of ``one-to-one-spawning'' selection mechanism, it resembles the Swarm Intelligence methods such as PSO but, at the same time it is similar to the Evolutionary Algorithms such as the GA, as it requires a ``mutation'' operator, followed by a ``croosover'' strategy to produce a new solution.

\InputIfFileExists{PSEUDOCODES/DE}{}{}

Despite the lack of proper fitness-informed selection and a fixed order for the operator, a number of different algorithmic behaviours can be obtained from DE by changing and combining mutation and crossover operators. The most commonly used mutation operators are:
\InputIfFileExists{MISC/DEMutationOperators}{}{}

Other configurations of DE exist and are described in the literature \cite{bib:DEbook}. With reference to equations (\ref{eq:derand1}) to (\ref{eq:derand2}), indices $r_1\neq r_2\neq r_3$ are randomly sampled in order to pick the random individual from the populations.

In modern terminology, the initial version of DE proposed in 1995 is DE/rand/1/bin; it includes a ``binomial'' crossover where each variable has a prefixed probability $CR$ to be exchanged. In subsequent implementations, a new crossover strategy has been proposed which exchanges bursts of consecutive components whose length depends on the $CR$ value and the dimensionality of the problem. As the probability of exchanging one more coordinate in the burst follows the geometric progression, and thus it decays exponentially, this variant is commonly referred to as DE/rand/1/exp. Implementation details of bin and exp crossover strategies are given in Algorithm \ref{alg:xobin} and \ref{alg:xoexp} respectively.
\InputIfFileExists{PSEUDOCODES/bin}{}{}
\InputIfFileExists{PSEUDOCODES/exp}{}{}

In modern terminology, any particular scheme can be obtained by combining the chosen mutation and crossover strategies which is reflected in the ``DE/a/b/c'' notation. The first member, ``a'', refers to the vector being mutated (namely the one to which difference vectors are added), ``b'' is the number of difference vectors used and ``c'' indicates the crossover. For example, ``a'' could be ``rand'' (Formulas \ref{eq:derand1} and \ref{eq:derand2}), ``best'' (Formulas \ref{eq:debest1}) or even a combination of two vectors as in ``current-to-best'' (Formula \ref{eq:decurtobest}). 


Listed mutations provide a set of different moves across the search domain for handling different landscape scenarios. The current-to-best/1 strategy, for instance, can be of help in speeding up the convergence when dealing with less complex fitness functions (functions with plateau-like regions as opposed to highly multimodal or ill-conditioned functions) \cite{bib:rotationEvo,rot2018}. However, it could be inadequate for highly multimodal functions, since it gives privilege to the direction towards the current best solution -- a basic rand/1 would be preferred in this case.

Despite the low number of parameters in DE, their tuning plays a major role as performances can heavily depend on the parameter setting. A skilled algorithm designer may exploit these while designing hybrid structures, as the adequate combination of the scale factor and crossover rate can theoretically be used to control the kind of desired search, hence the ``control parameters'' name, so to reproduce either the more exploitative behaviours or the more local-search-like routine, simply by fine picking $F$ and $CR$. Unfortunately, this is not always achievable due to the lack of precise indications. The scale factor $F$, which is used to control the exploitative lengths of mutations operators, has been initially thought to have a large range, i.e. $F\in(0,2]$, where the the value $F=0$ is usually not considered as it will nullify the effect of the difference vector of the mutation operator (see Formulas \ref{eq:derand1} to \ref{eq:decurtobest}). On the contrary, the crossover rate has been expected to lie in the range $[0,1]$, probabilistically, where extreme values are typically avoided to prevent the crossover operator to either replace only $1$ component(see Algorithm \ref{alg:xobin} and \ref{alg:xoexp}) or to return an exact copy of one of the parents. However, later on it has become more clear that only values of $F$ in $(0,1]$ were used in practical applications. Moreover, the subsequent study in \cite{tuningLampinen} recommended $F\in[0.5,0.9]$ and $CR\in [0.8,1]$. In practice, users of DEs tend to settle on values close to $0.7$ for the scale factor and several DE algorithms have been proposed in which such parameters are self-adapted to the problem during the optimisation process \cite{bib:Brest2006a}, or picked up from a pool of promising values \cite{cheng2015}. A key study in \cite{bib:Zaharie2002} has instead dug into the interrelation among parameters $F$ and $CR$ and discovered an optimal tuning of the crossover rate in function of the scale factor.


Therefore, while research literature is rich of studies on setting $F$ and $CR$, significantly less information is available on setting the value of $NP$ and some studies are also arguable. For example, \cite{tuningLampinen} recommended to use a population made up of $10$ times the value of dimensionality of the problem. Such recommendation appears to be unfeasible in many real-world and large-scale scenarios \cite{bib:Parsopoulos2009}, and, more importantly, it directly contradicts the observation that shrinking the population size can be beneficial to avoid stagnation \cite{bib:Brest2008b}. 

Furthermore, large population size is known to promote diversity, thus reducing the risk of premature convergence \cite{bib:yaman2018}. However, a too large population size could potentially unnecessarily strengthen the deleterious effect of the structural bias. As demonstrated theoretically in \cite{KONONOVA2015} in case of GAs, the spread of the population across the domain is directly effected by the number of points in the population: unfortunately enough, as population size grows, so does the strength of structural bias. However, the situation is less clear for other optimisation frameworks and, in particular, for DE. According to the preliminary study in \cite{bib:biasDELego18} such correlation is only partially confirmed. Surprisingly, it turned out that while the ``current-to-best/1'' mutation operator seems to carry a visible structural bias, a simpler ``rand/1'' is to some extent capable of mitigating the bias regardless of the chosen crossover strategy. Thus, \cite{bib:biasDELego18} is extended here for a wider range of of aspects such as different parameter settings, corrections strategies and DE schemes.

\section{Objectives, methods and experimental setup}\label{OandM}
The main objective of this publication is to analyses a wide range of popular DE configurations and identify a combination of mutation and crossover operators responsible for the rise of the structural bias in DE algorithms, as well as to observe to which extent the choice for the control parameters $F$, $CR$, correction strategy and population size $NP$, can mitigate such bias. This knowledge can be exploited in the long term perspective to achieve a more informed algorithmic design process and to give practitioners guidance in tuning DE's control parameters for real-world optimisation problems.

\subsection{Choice of DE configurations}
As explained in Section \ref{testingForBias}, identification of structural bias is based on the minimisation of function (\ref{eq:f0}) \footnote{without loss of generality, conclusions here also hold true for the maximisation case}. For historical reasons, dimensionality is kept at $n=30$. To leave nothing to chance, the effect of different correction strategies is also taken into consideration in this study. Experiments in this article have been carried out with the following implementations of popular correction methods introduced in Section \ref{constrained}:
\begin{enumerate}
\item \textbf{penalise}:  $f_0:[0,1]^n\to [0,1]$ is extended via a surjective penalty function $f_P:\mathbb{R}^n \to [0,1]\cup\{c\}$ mapping solutions outside the $[0,1]^n$ domain with a pre-fixed real valued constant $c\not\in[0,1]$:
\begin{equation}\label{eq:penalty}
f_{P}(\mathbf{x})=\left\{
\begin{array}{lr}
      f_0 & \text{if }\mathbf{x}\in [0,1]^n \\
      2 & \text{otherwise} 
\end{array}\text{;} \right.
\end{equation}
\item \textbf{saturation}: keeping the original fitness value, modify those coordinates outside the domain as shown in Figure \ref{fig:sketch_sat} and  Algorithm \ref{alg:saturation} (superficial correction);
\item \textbf{toroidal}: keeping the original fitness value, modify those coordinates outside the domain as shown in Figure \ref{fig:sketch_tor} and Algorithm \ref{alg:toro} (superficial correction).
\end{enumerate}

It must be pointed out that a popular ``\textit{dismiss}'' correction strategy is indirectly omitted from this study as it would be equivalent to using a penalty function, due to the one-to-one spawning logic in DE. Indeed, in the most general case this is implemented as in Algorithm \ref{alg:discard}, in which the unfeasible solution is simply discarded and replaced with a parent chosen according to any convenient logic. However, the same replacement necessarily occurs in DE for penalised solutions as they are discarded based on a direct comparison on their fitness function value. Differently, it makes sense to differentiate between ``penalise'' and ``dismiss'' methods in EA paradigms such as e.g. some GAs for which selection takes place stochastic.
\InputIfFileExists{PSEUDOCODES/saturation}{}{}

\InputIfFileExists{PSEUDOCODES/toroidal}{}{}
\InputIfFileExists{PSEUDOCODES/discard}{}{}

The mutation strategies presented in Section \ref{DE}, Formulas \ref{eq:derand1} to \ref{eq:decurtobest}, were combined with both bin and exp crossovers, i.e. Algorithms \ref{alg:xobin} and \ref{alg:xoexp} respectively, thus, forming $8$ DE schemes. Each one was equipped with the $3$ aforementioned correction schemes (i.e. Formula \ref{eq:penalty}, Algorithm \ref{alg:saturation} and \ref{alg:toro}) thus generating $24$ different DE configuration variants. 

\subsection{Choice of DE parameters}
To decide on the most appropriate parameters settings for DE variants selected in the previous section, an extensive preliminary experimentation has been carried out. Keeping the number of runs and population sizes consistent with  \cite{KONONOVA2015,bib:biasDELego18}, DE control parameters were \textit{investigated in the light of the percentage of points corrected throughout the whole optimisation process}. The logic behind such investigation is to focus on the algorithmic behaviour and stick as much as possible to optimising the true $f_0$ thus minimising potential side effects introduced by the correction strategy. In this sense, percentage of corrections during the run is \textit{a metric} of how aggressive, or badly-suited, the correction strategy is. On the other hand, the choice of control parameters of DE is balanced empirically by the fact that too few corrections potentially mean underexploitation of the domain boundaries as generating operators have relatively local nature. Thus, the most reliable $F$-$CR$ pair was determined by looking into $25$ pre-selected combinations of control parameters $F$ and $CR$, i.e. those generated with $F \in \lbrace 0.05, 0.2, 0.4, 0.7, 0.9\rbrace$ and $CR\in \lbrace 0.05, 0.4, 0.7, 0.9, 0.99\rbrace$. Such tabulation of the intervals of control parameters is empirically motivated by regions of particular interest.  


This means that $24$ configurations of DE have been considered with $3$ population sizes each producing $25$ histograms (one per each $F-Cr$ couple) -- resulting in $1800$ full optimisation runs being performed each $50$ times to keep track of the percentage of occurred corrections. From distributions of these percentages of corrections, average and standard deviation surface plots have been produced to visually handle such amount of data and values of $0.1$-$0.2$ of the $F-Cr$ control parameter couple have been chosen to force the algorithms under study to operate, as much as possible, within the problem's boundaries. To avoid an unnecessarily long list of graphs here, all these results are made available online \cite{bib:BiasDEResults}. 

\begin{figure}[H] \centering
\subfigure[{Penalty correction}, ${NP{=}5}$.]{\includegraphics[width=.32\linewidth,angle=180]{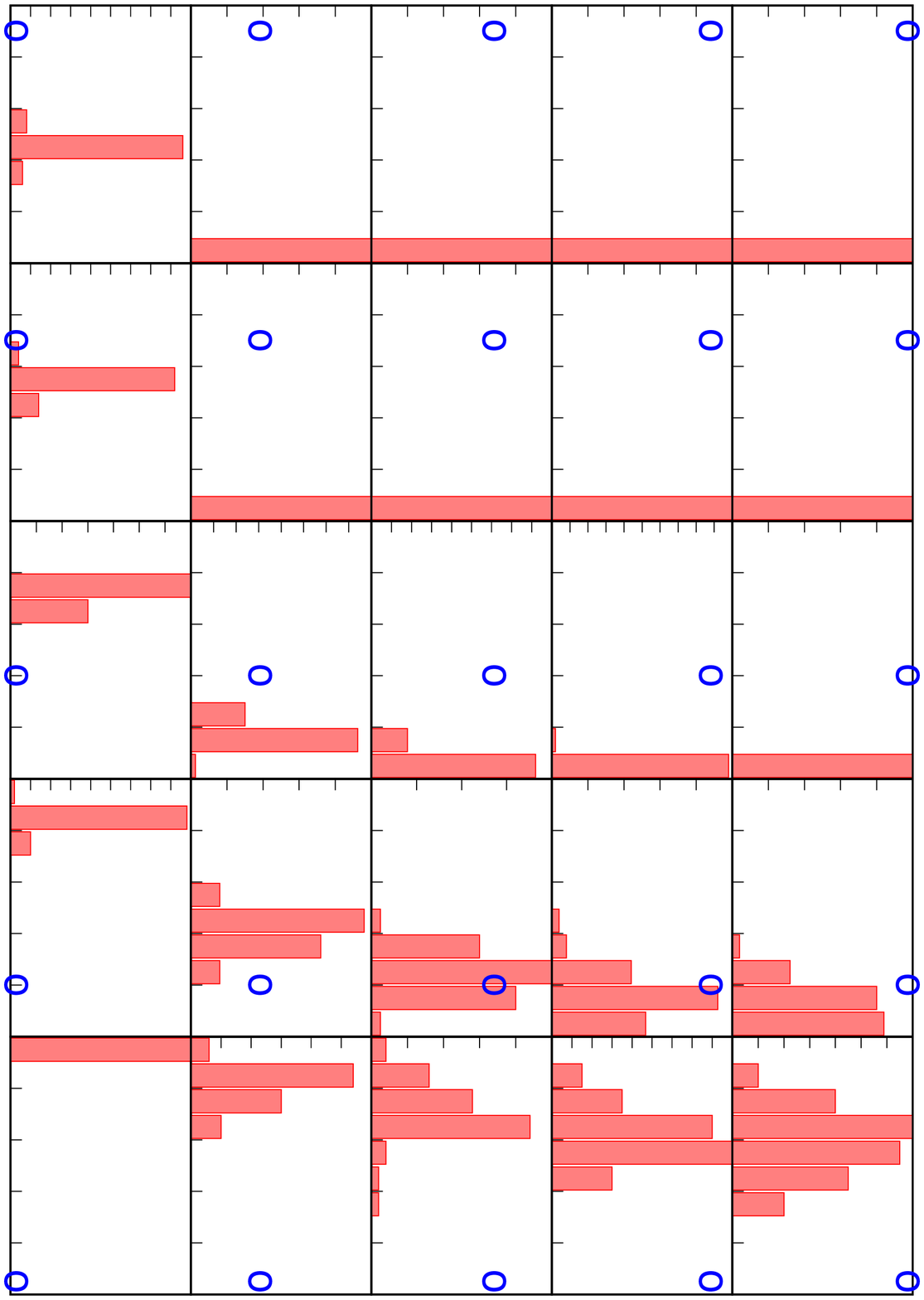}}
\subfigure[{Penalty corection}, ${NP{=}20}$.]{\includegraphics[width=.32\linewidth,angle=180]{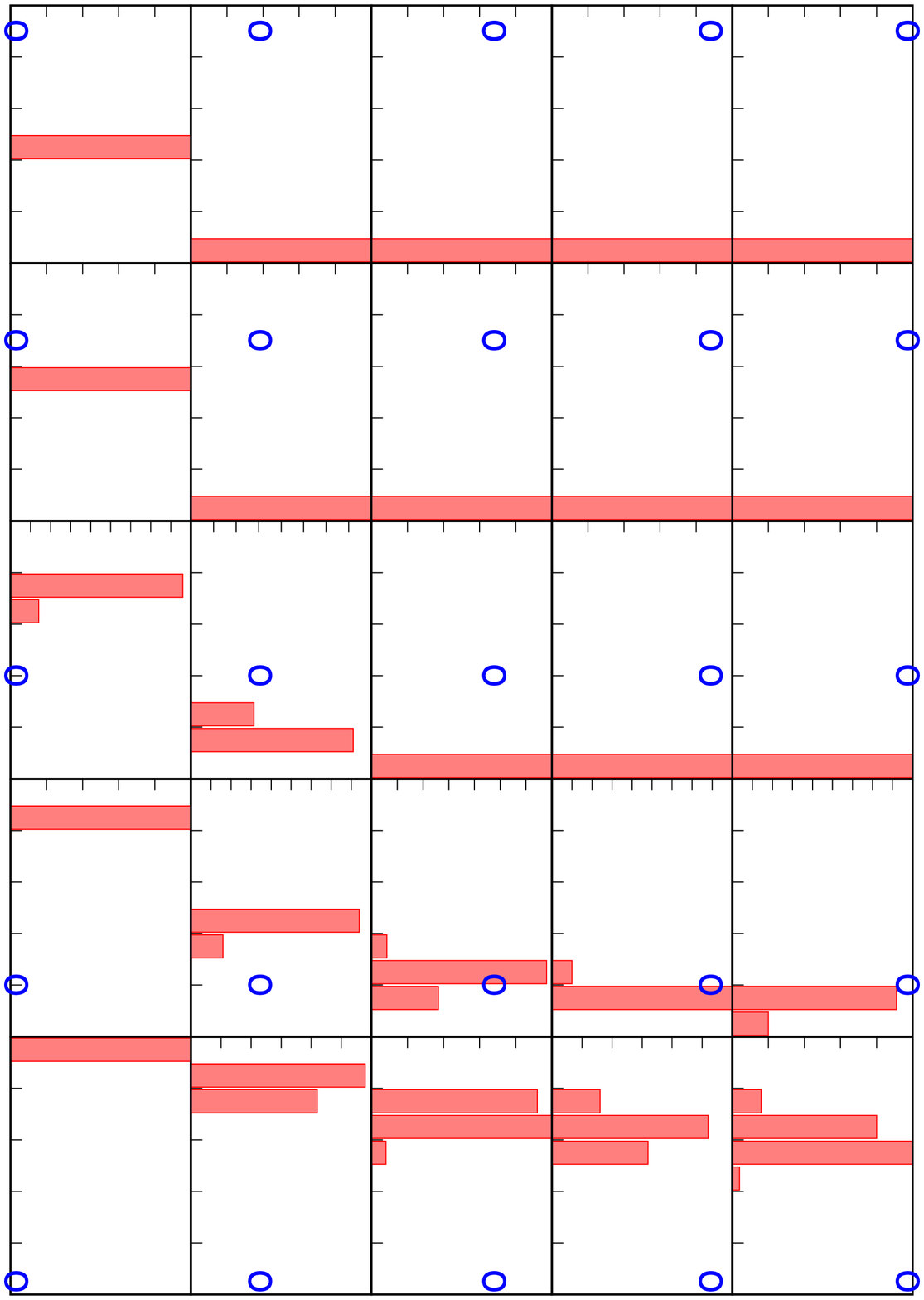}}
\subfigure[{Penalty corection}, ${NP{=}100}$.]{\includegraphics[width=.32\linewidth,angle=180]{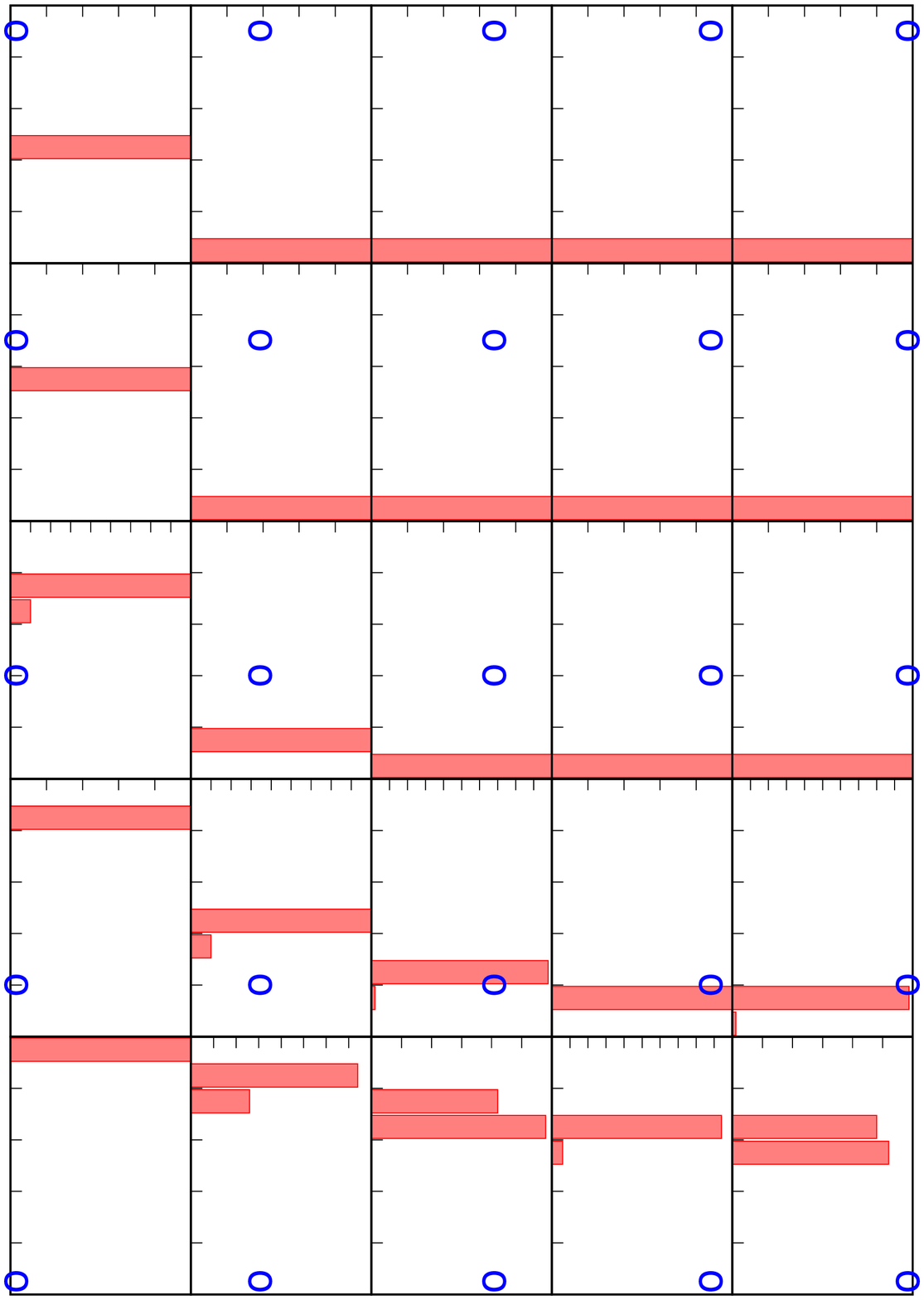}}
\caption{Example of distributions of percentage of corrections for {DE/rand/1/bin} for various values of $F$ and $CR$. For explanation of axes and colours, see the text.}\label{fig:disDEROB}
\end{figure}

For demonstration purposes only, one example for the DE/rand/1/bin with penalty correction case and $3$ different population sizes is shown here in Figure \ref{fig:disDEROB}. Detailed analysis of the percentage of corrections in different configurations of DE has proved to be very interesting and will be published shortly as a separate publication. To clarify, Figure \ref{fig:disDEROB} depicts distributions of the percentage of occurred corrections, while optimising $f_0$, for each combination $F$ and $CR$ (i.e. $25$ sub-diagrams per subfigure). Each sub-diagram carries two layers of information. The first layer, shown in red, represents the distribution of correction percentages in a series of $50$ runs. Percentages can be read on the y-axis (which points upwards and whose range is always $[0,1]$), while the number of runs is reported on x-axis (which points to the right). The second layer, shown in blue, indicates the employed values of $F$ and $CR$. Also for this layer, the y-axis points upwards and the x-xis to the right, but they report values for $F$ and $CR$ respectively. For both layers, origin is in the lower left corner. This Figure attempts to draw a three-dimensional figure in projections in two dimensions: distributions shown in red in each small subfigure should be placed on the page in a perpendicular fashion towards the reader, in points marked in blue circles. Choice of number of bins in histograms is based on the investigation explained in the next section.

To summarise, having chosen appropriate values of DE control parameters, for the purpose of studying structural bias in DE configurations, all aforementioned $24$ DE configuration variants were each run $50$ times on $f_0$ for the $n=30$ dimensional case with a standard computational budget of $10000\times n=300000$ fitness functional calls with $F=0.1$, $Cr=0.2$ and population size of $NP\in\lbrace 5,20,100\rbrace$. Thus, in total, $24\times3=72$ optimisation processes were repeated for $50$ runs to generate the results shown in Section \ref{results}.

\subsection{Choice of granularity}
\begin{figure}[H] \centering
\subfigure[{5 bins}, ${NP{=}5}$.]{\includegraphics[width=.32\linewidth,angle=180]{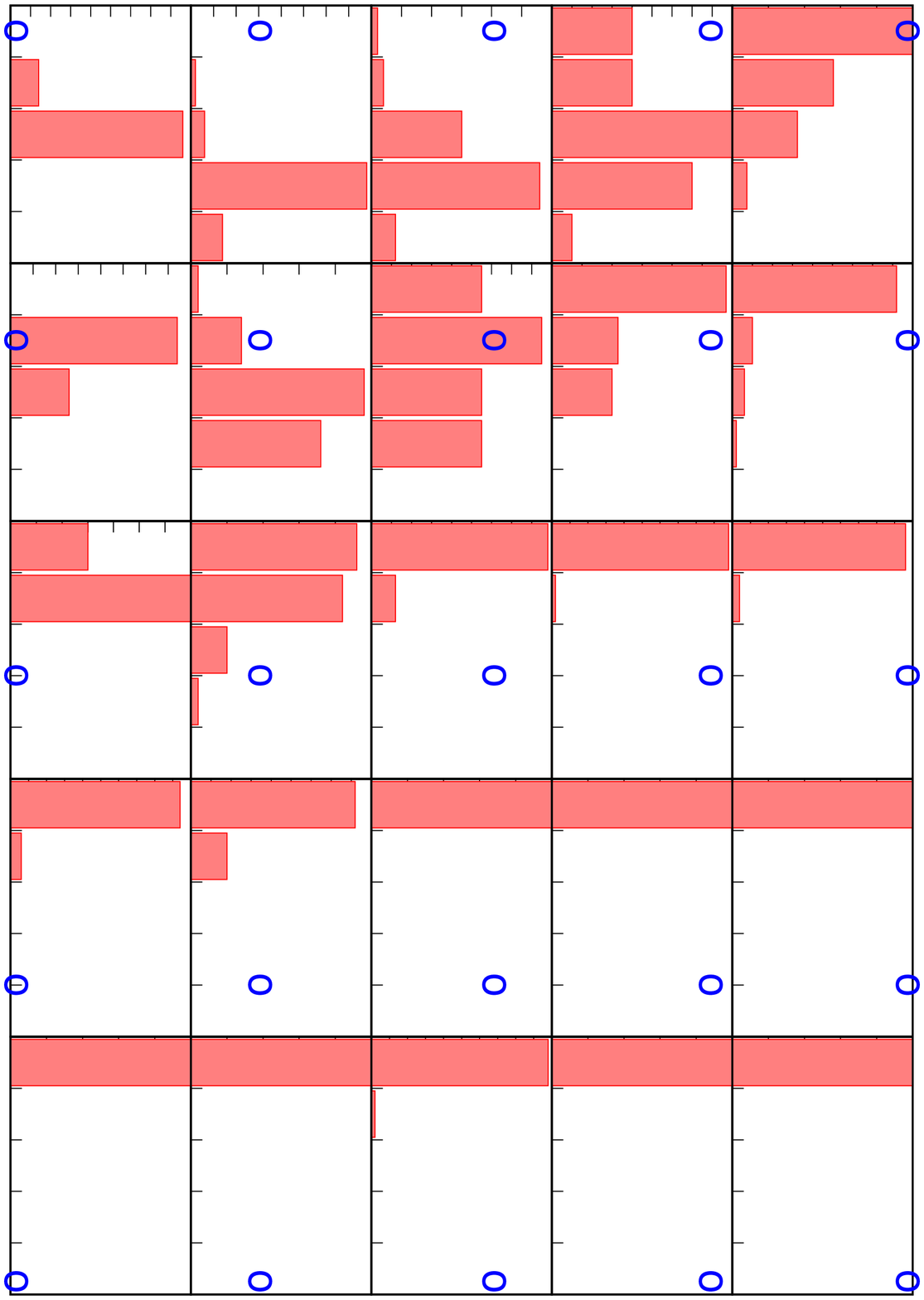}} 
\subfigure[{5 bins}, ${NP{=}20}$.]{\includegraphics[width=.32\linewidth,angle=180]{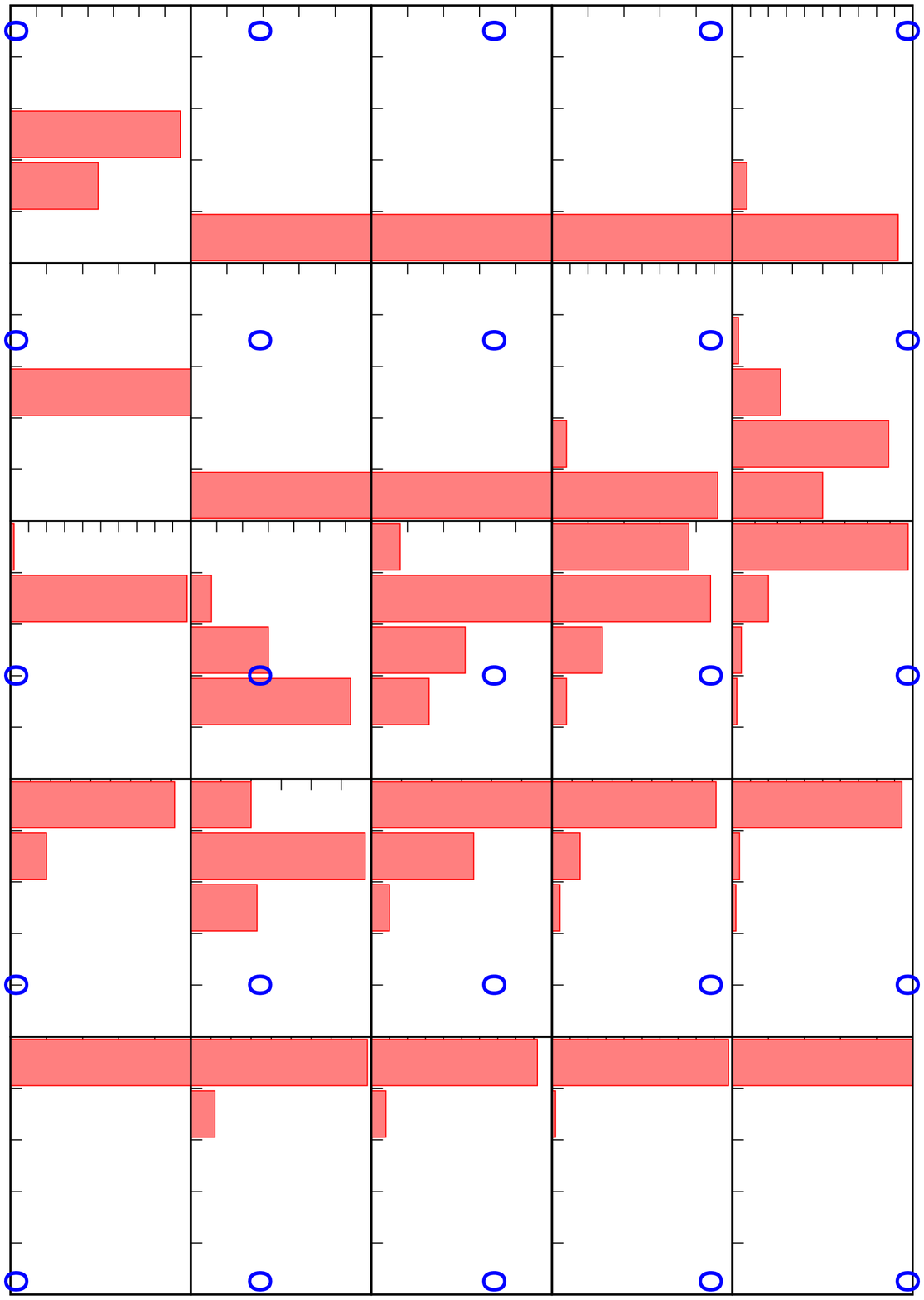}} 
\subfigure[{5 bins}, ${NP{=}100}$.]{\includegraphics[width=.32\linewidth,angle=180]{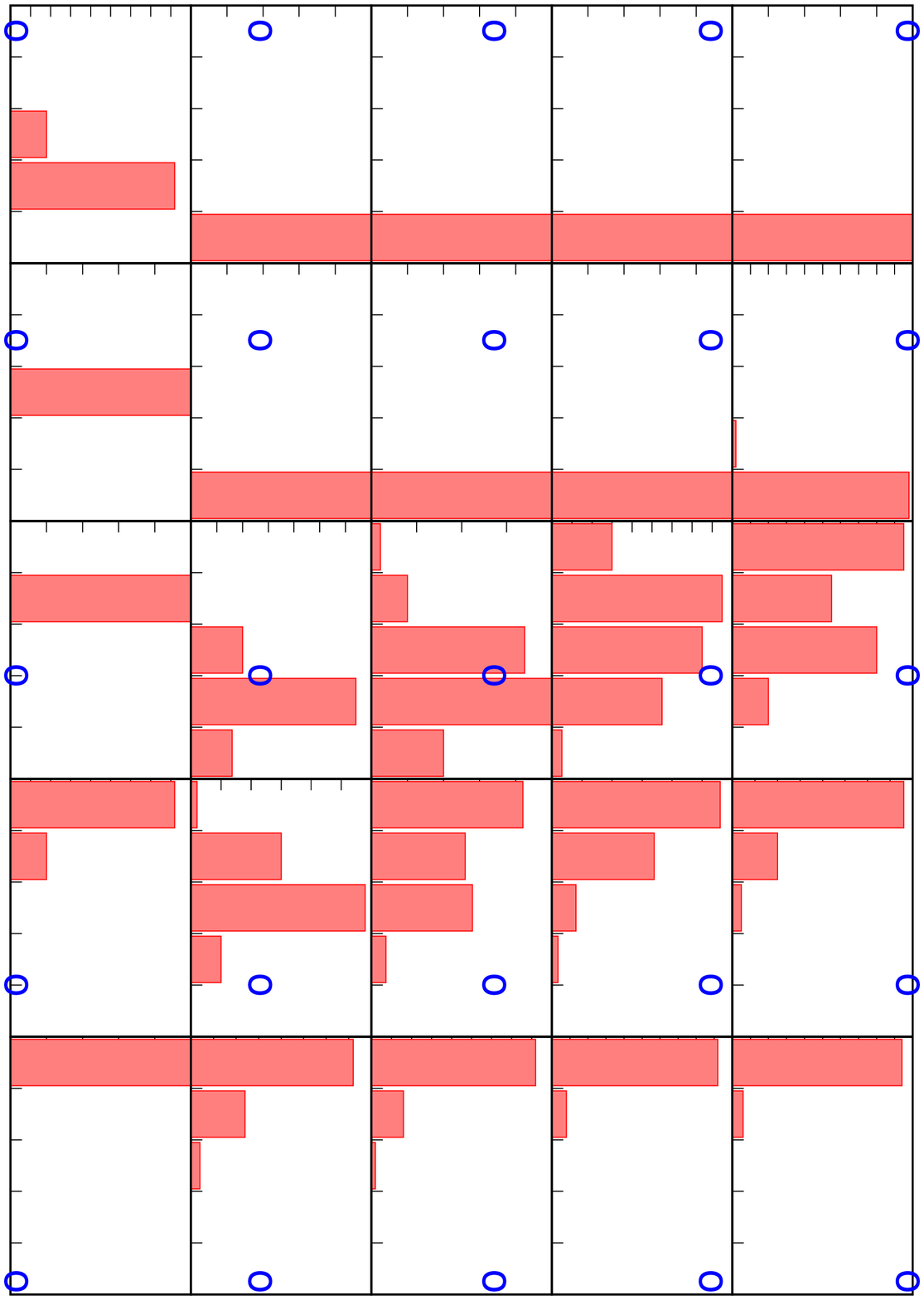}} 
\subfigure[{25 bins}, ${NP{=}5}$.]{\includegraphics[width=.32\linewidth,angle=180]{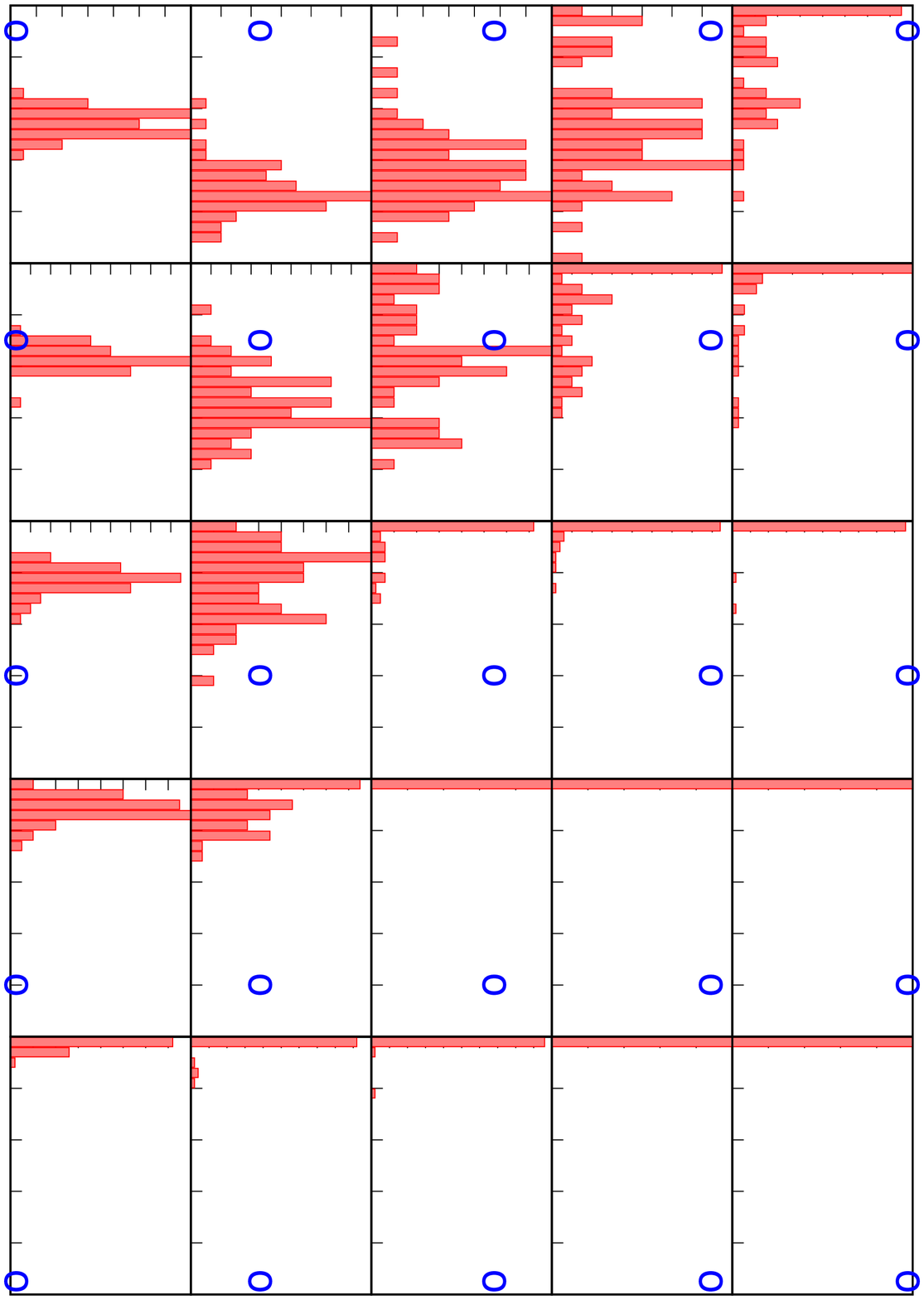}} 
\subfigure[{25 bins}, ${NP{=}20}$.]{\includegraphics[width=.32\linewidth,angle=180]{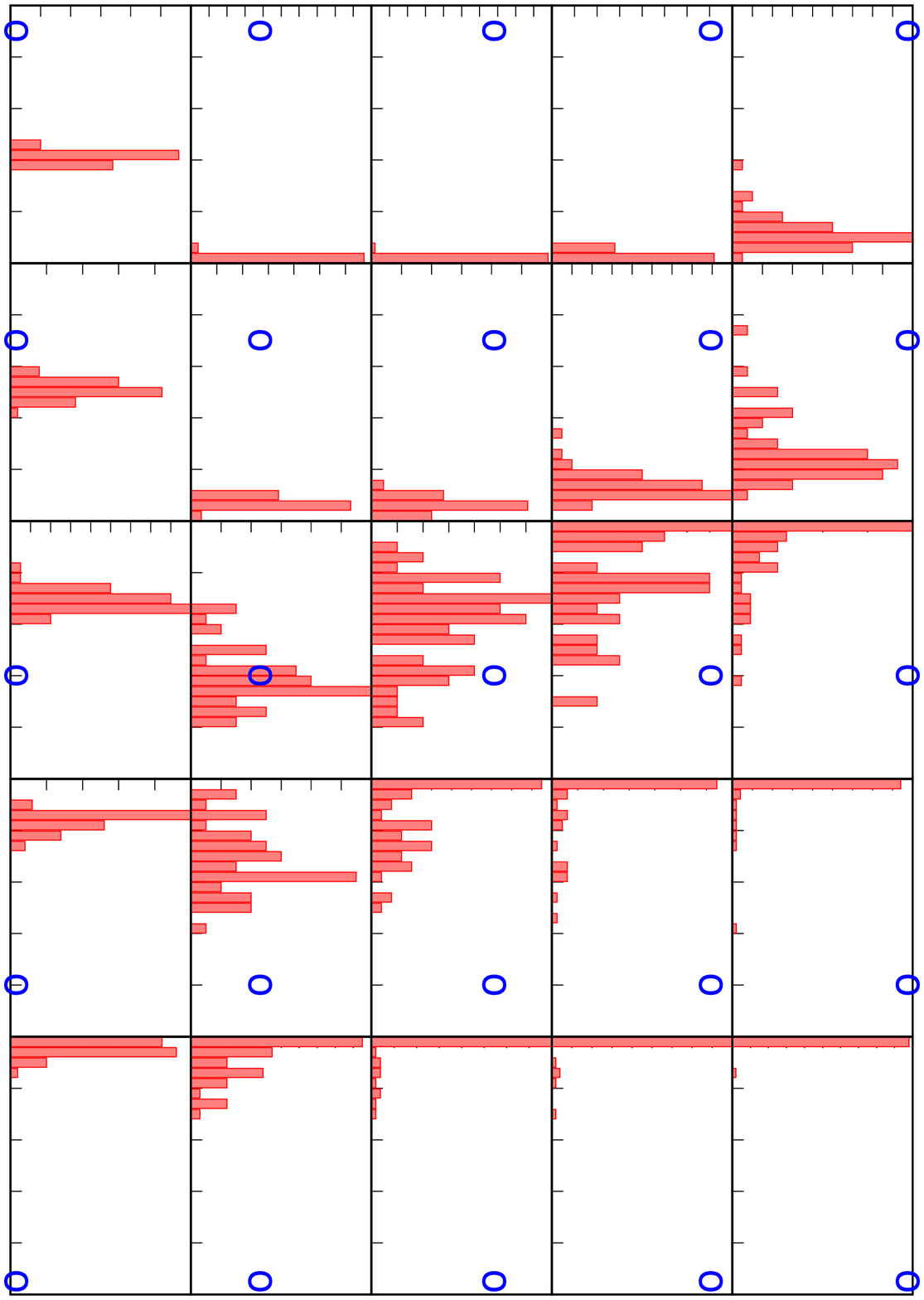}} 
\subfigure[{25 bins}, ${NP{=}100}$.]{\includegraphics[width=.32\linewidth,angle=180]{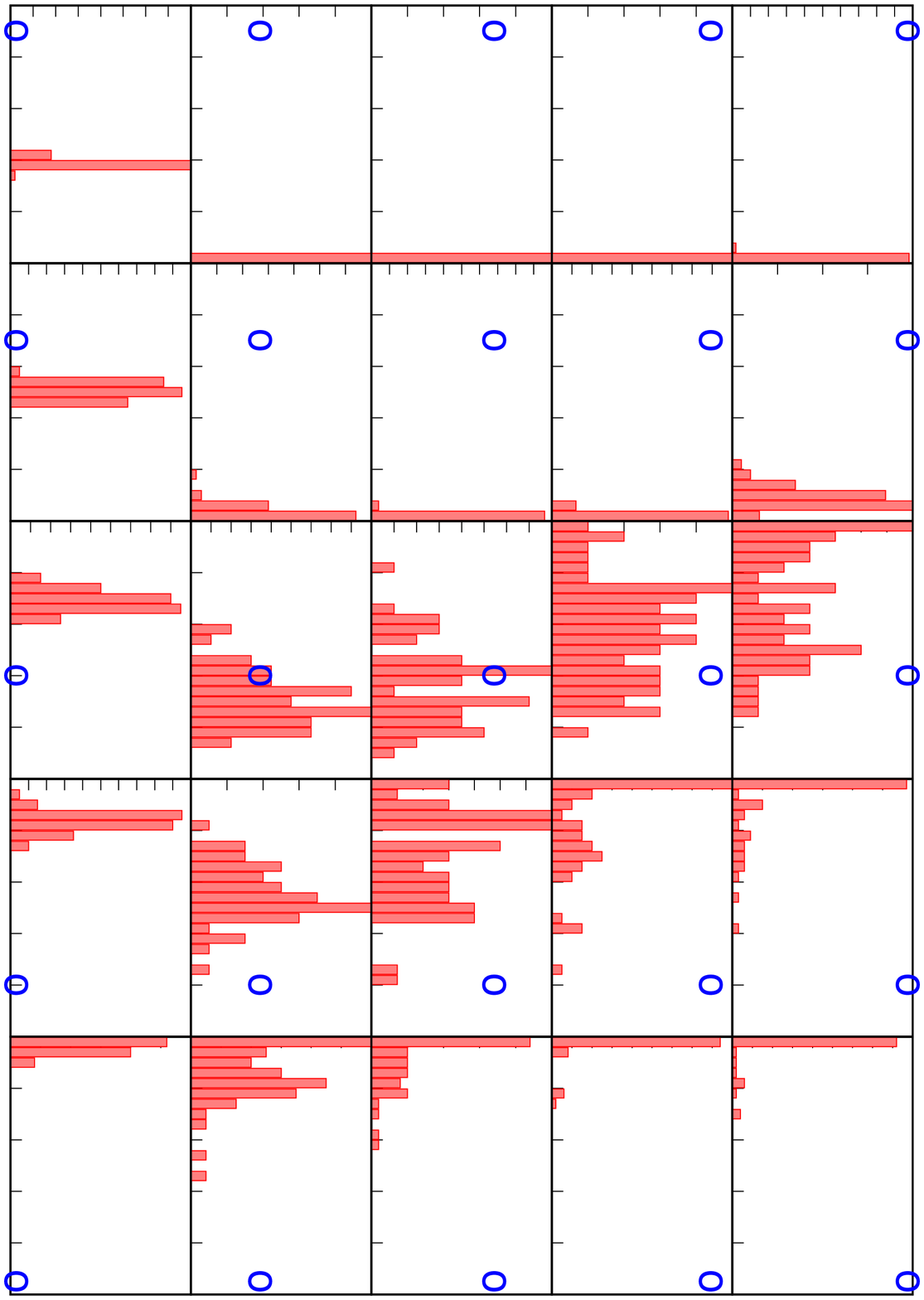}} 
\subfigure[{50 bins}, ${NP{=}5}$.]{\includegraphics[width=.32\linewidth,angle=180]{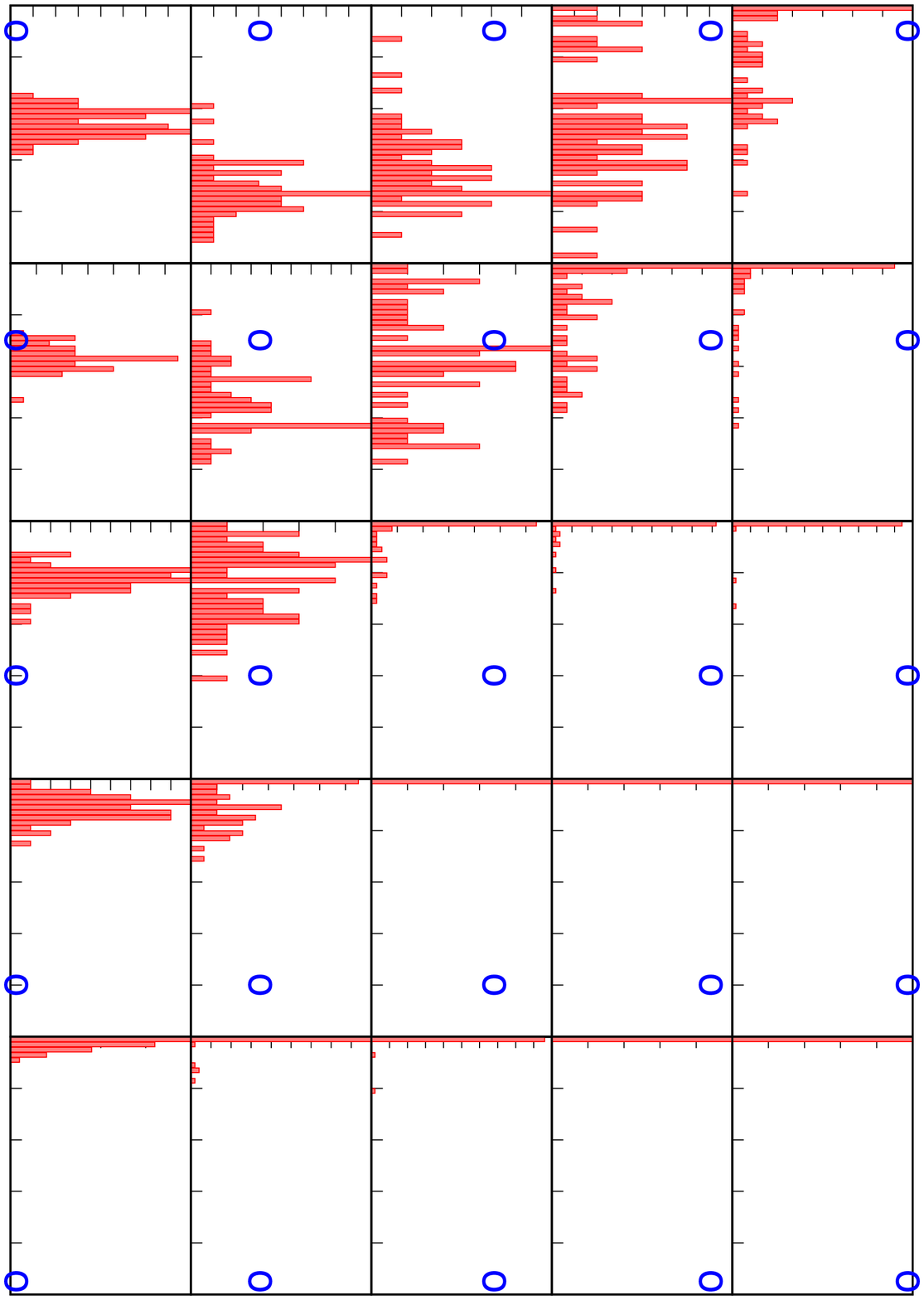}} 
\subfigure[{50 bins}, ${NP{=}20}$.]{\includegraphics[width=.32\linewidth,angle=180]{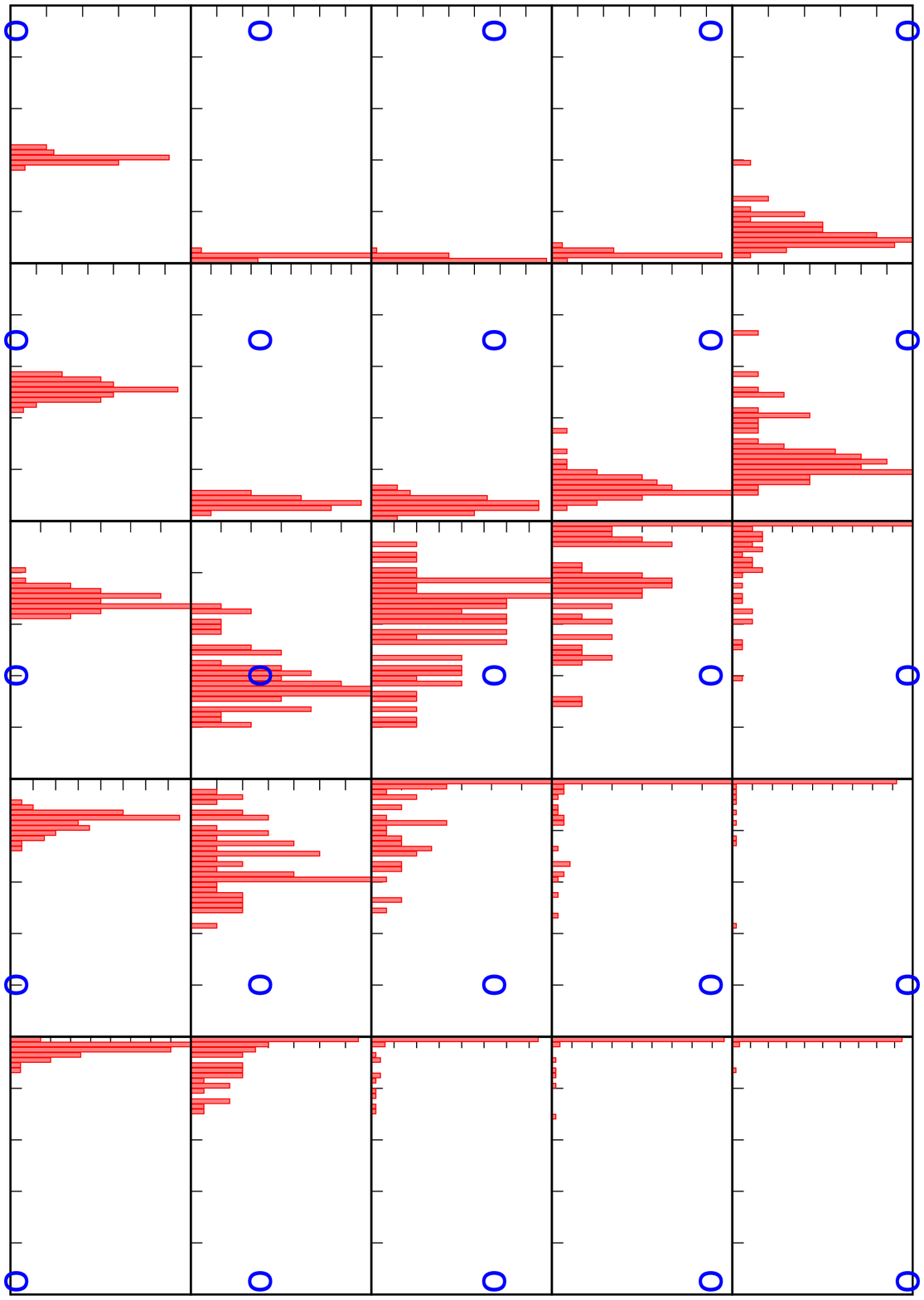}} 
\subfigure[{50 bins}, ${NP{=}100}$.]{\includegraphics[width=.32\linewidth,angle=180]{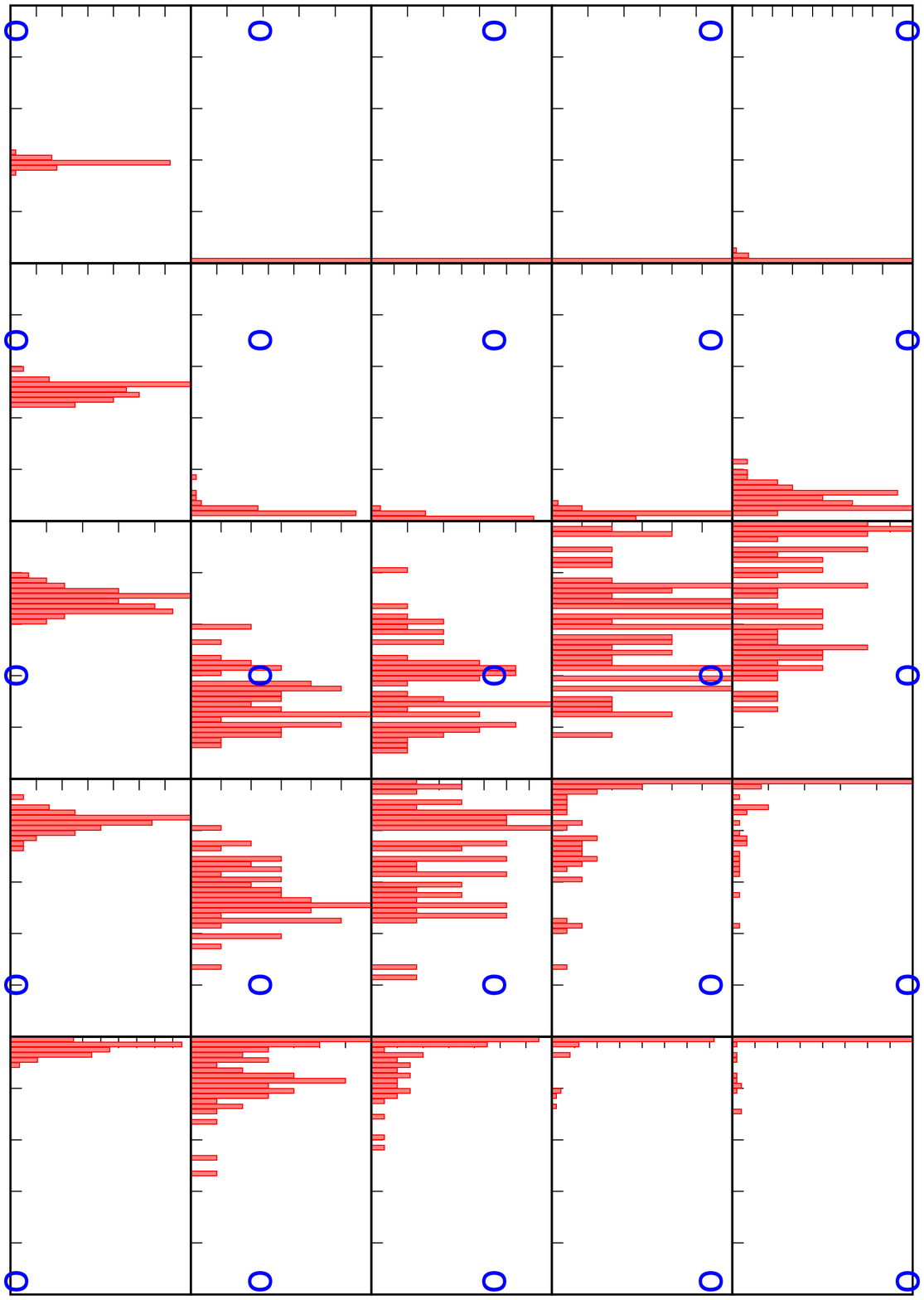}} 
\caption{Choosing the number of bins in distributions of the percentage of corrections for various values of $F$, $CR$ and populations sizes for \textbf{DE/best/1/bin} with saturation correction of $f_0$.}\label{fig:dis_bins}
\end{figure}

To generate meaningful distribution figures such as Figure \ref{fig:disDEROB}, it has become clear that bin size had to be tuned. Choosing a number of bins for a histogram is a balancing act which depends on the number of points in a histogram and features of the underlying true distribution of data (symmetry, skewness, etc.). Choosing a too small number of bins oversmooths the resulting histogram and potentially looses the distribution features. Meanwhile, a too high number of bins results in a too noisy picture which is difficult to interpret and generalise. Clearly, the number of bins should not exceed the number of data points. As no educated a priori assumptions could be made regarding properties of the distribution of the percentage of corrections, an initial study has been made to decide on the appropriate number of bins in distribution plots. DE/best/1/bin configuration with saturation correction has been chosen as representative and distributions of the percentage of corrections have been visualised for three population sizes for three different numbers of bins ($5$, $25$, $50$), see Fig. \ref{fig:dis_bins}. 

Clearly, histograms with $50$ bins (last row in Fig. \ref{fig:dis_bins}) are too noisy. Histograms with $25$ bins (middle row in Fig. \ref{fig:dis_bins}) barely show any qualitative differences with $50$ bins, thus they are too noisy as well. Histograms with $5$ bins (top row in Fig. \ref{fig:dis_bins}) are too crude. Thus, it has been decided to use $10$ bins for all histograms of the percentage of corrections as is done in Figure \ref{fig:disDEROB}. 

\subsection{Further results on distributions of the number of corrections}\label{sect:teaser}
\begin{figure}[tbh] \centering
\subfigure[{Penalty corection}, $NP=5$.]{\includegraphics[width=.23\linewidth,angle=270,trim={10mm 20mm 10mm 10mm},clip]{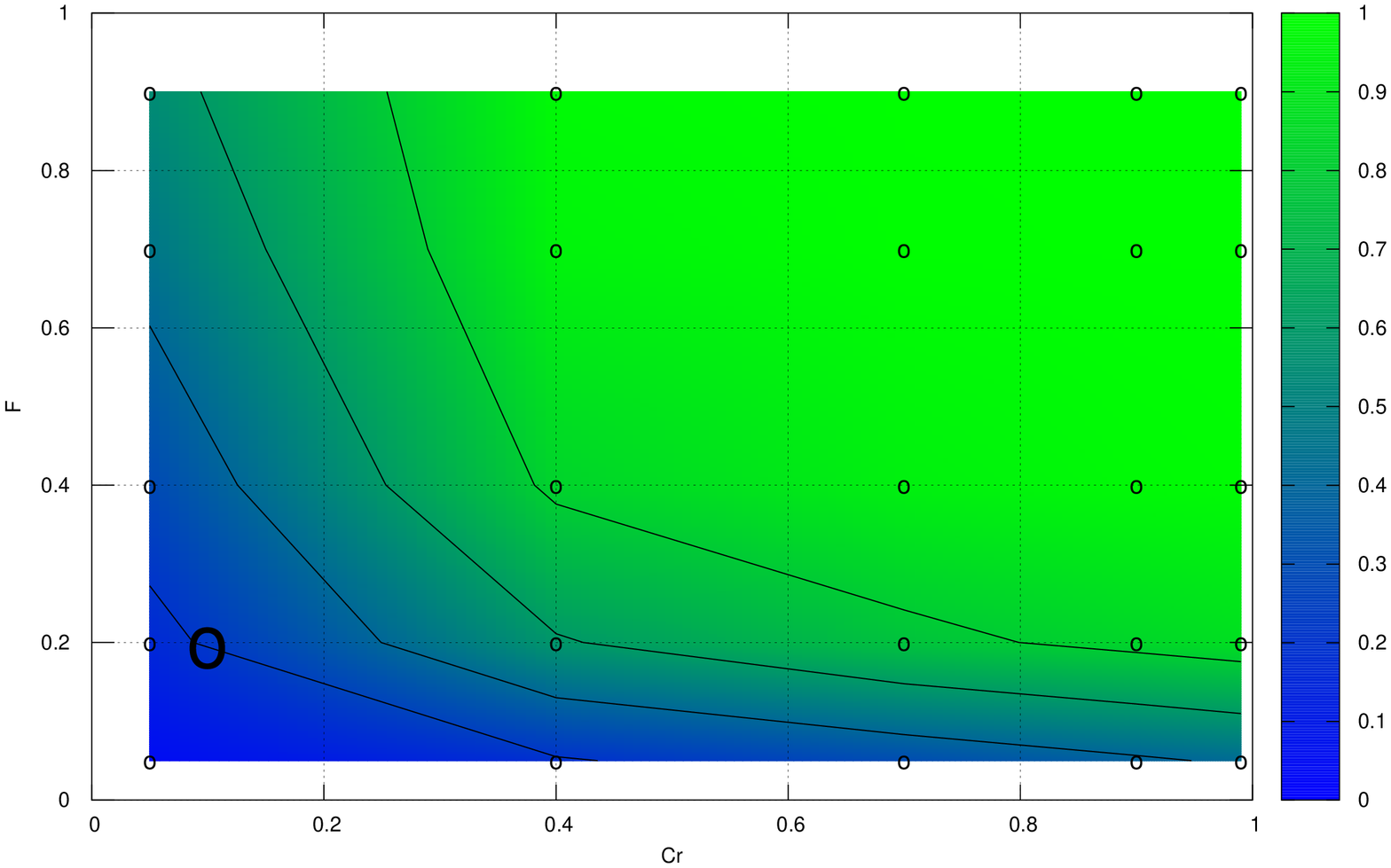}}
\subfigure[{Penalty corection}, ${NP\boldsymbol{=}20}$.]{\includegraphics[width=.23\linewidth,angle=270,trim={10mm 20mm 10mm 10mm},clip]{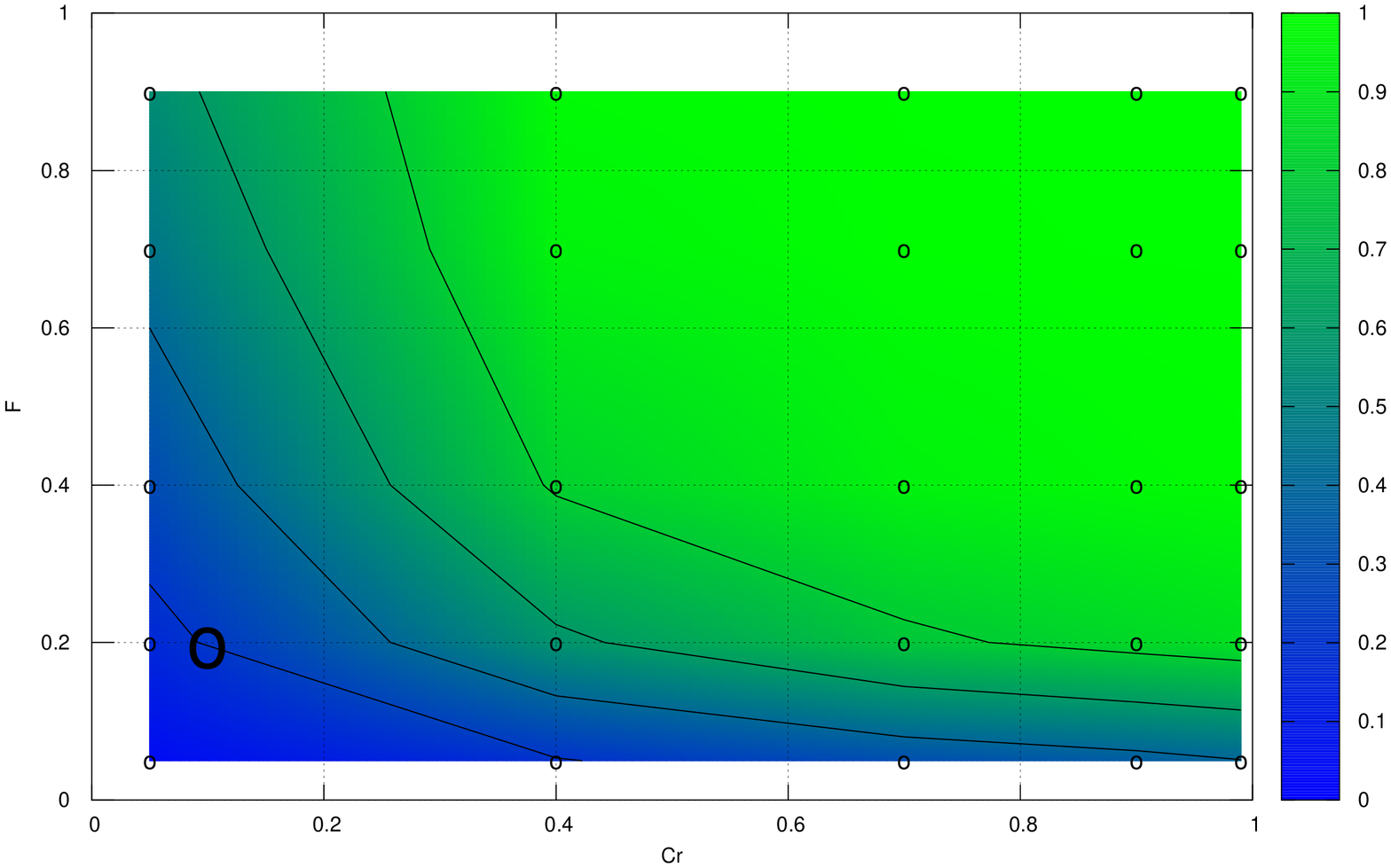}}
\subfigure[{Penalty corection}, ${NP\boldsymbol{=}100}$.]{\includegraphics[width=.23\linewidth,angle=270,trim={10mm 20mm 10mm 10mm},clip]{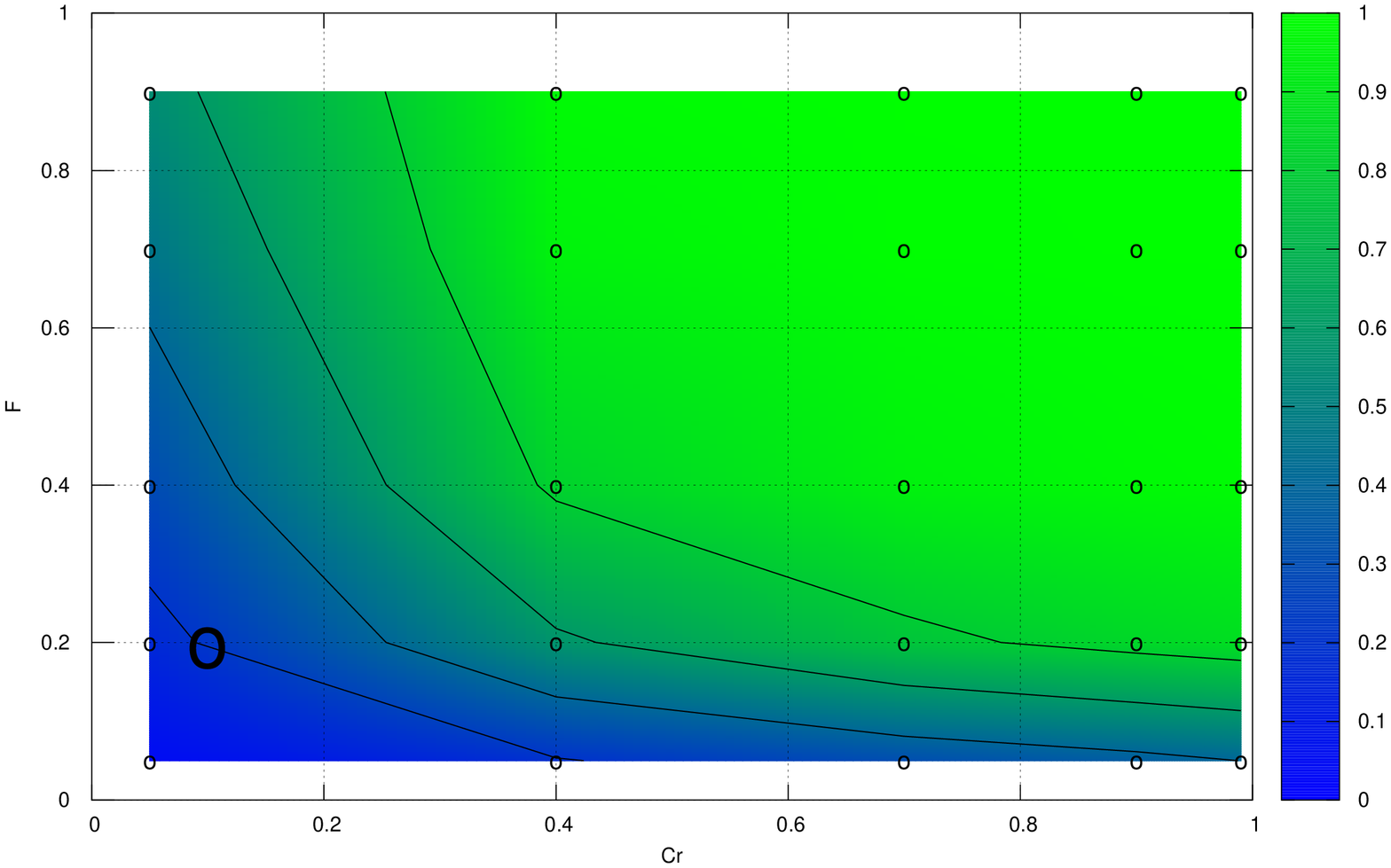}}
\caption{Example of interpolated surface of average percentage of corrections for {DE/rand/1/bin} for various values of $F$ and $CR$. For explanation about axes, see Section \ref{sect:teaser}.}\label{fig:surDEROB}
\end{figure}

To help with visualising multidimensional data, Figures \ref{fig:surDEROB} depicts the average percentage of corrections for DE/rand/1/bin configuration, over $50$ runs, for three population sizes, occurring for each one of the $25$ $F-Cr$ combination. More specifically, in each subfigure, axes pointing upwards shows values of $F$, meanwhile axis pointing to the right shows values of $CR$. Colour encodes the value of the a posteriori computed average percentage of corrections for experiments with given $F$ and $CR$ values -- experimentally obtained values of averages are marked with small dots, meanwhile averages for the rest of $F-Cr$ couples are interpolated and smoothed. Black curves are interpolated contour curves for the average percentage of corrections. Colour axis is the same for all subfigures -- it runs from $0$ values in blue to $1$ values in green. Chosen $F-Cr$ paid is marked with a small circle. 

\begin{figure}[H] \centering
\subfigure[{Penalty corection}, ${NP{=}5}$.]{\includegraphics[width=.23\linewidth,angle=270,trim={10mm 20mm 10mm 10mm},clip]{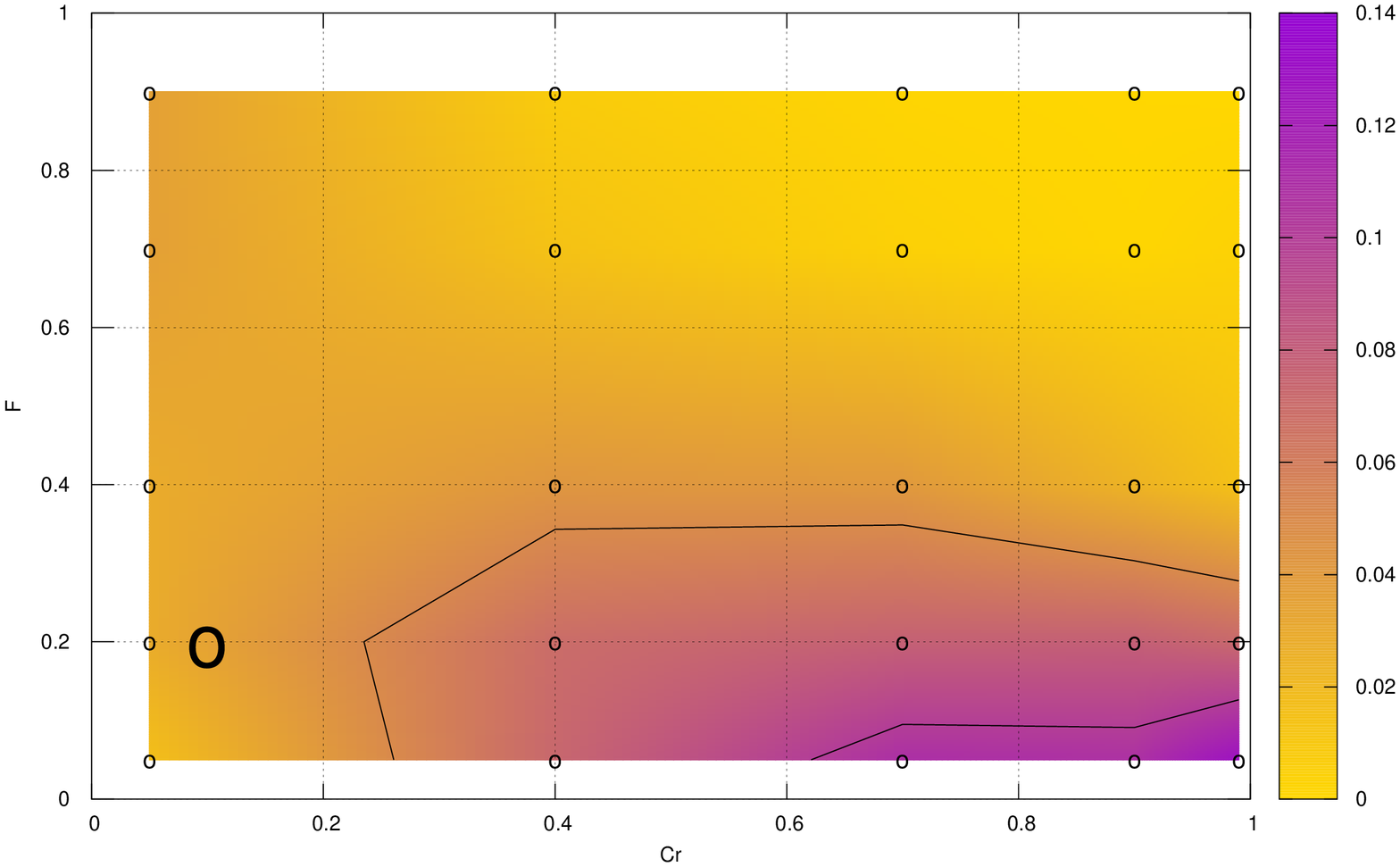}}
\subfigure[{Penalty corection}, ${NP{=}20}$.]{\includegraphics[width=.23\linewidth,angle=270,trim={10mm 20mm 10mm 10mm},clip]{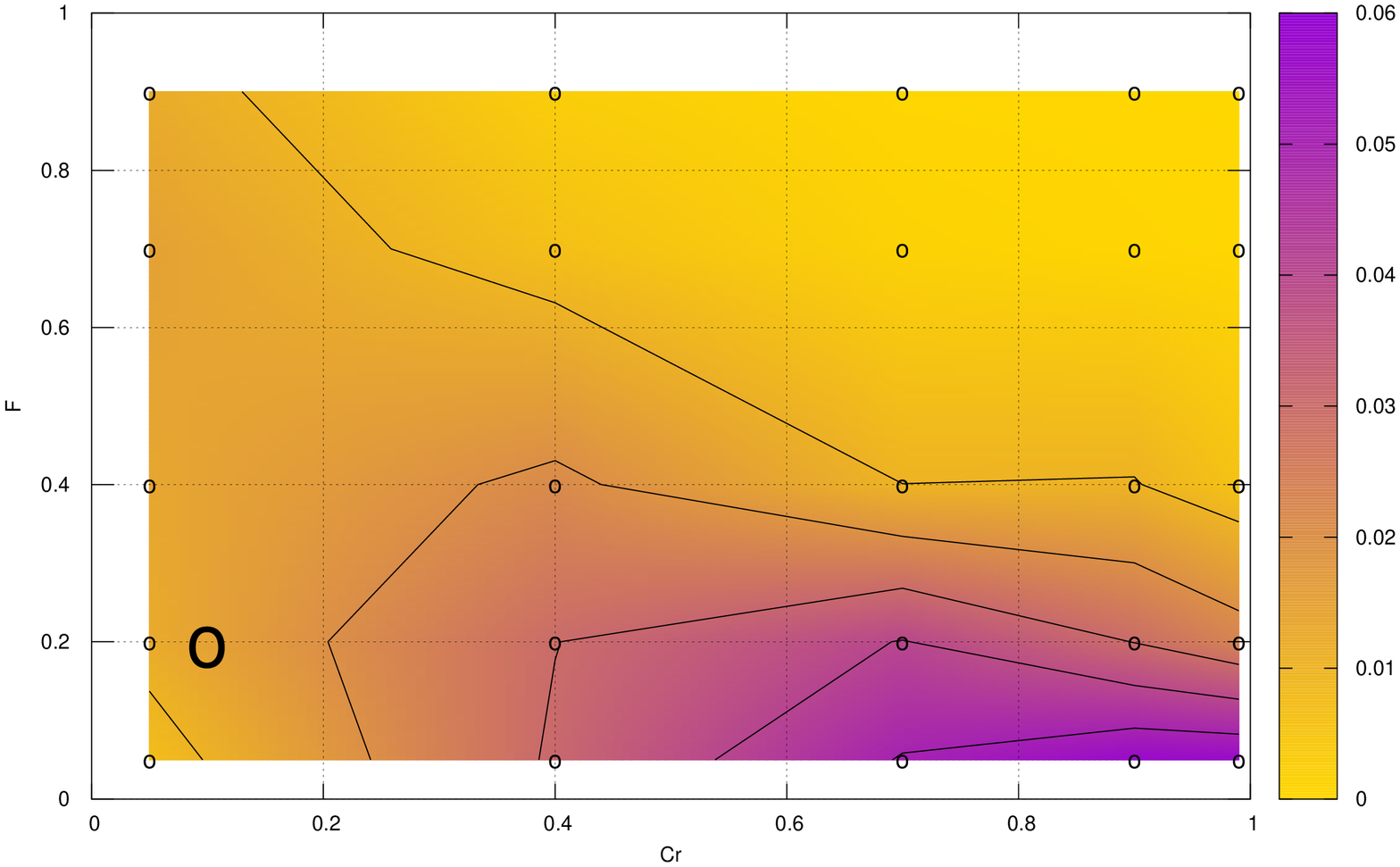}}
\subfigure[{Penalty corection}, ${NP{=}100}$.]{\includegraphics[width=.23\linewidth,angle=270,trim={10mm 20mm 10mm 10mm},clip]{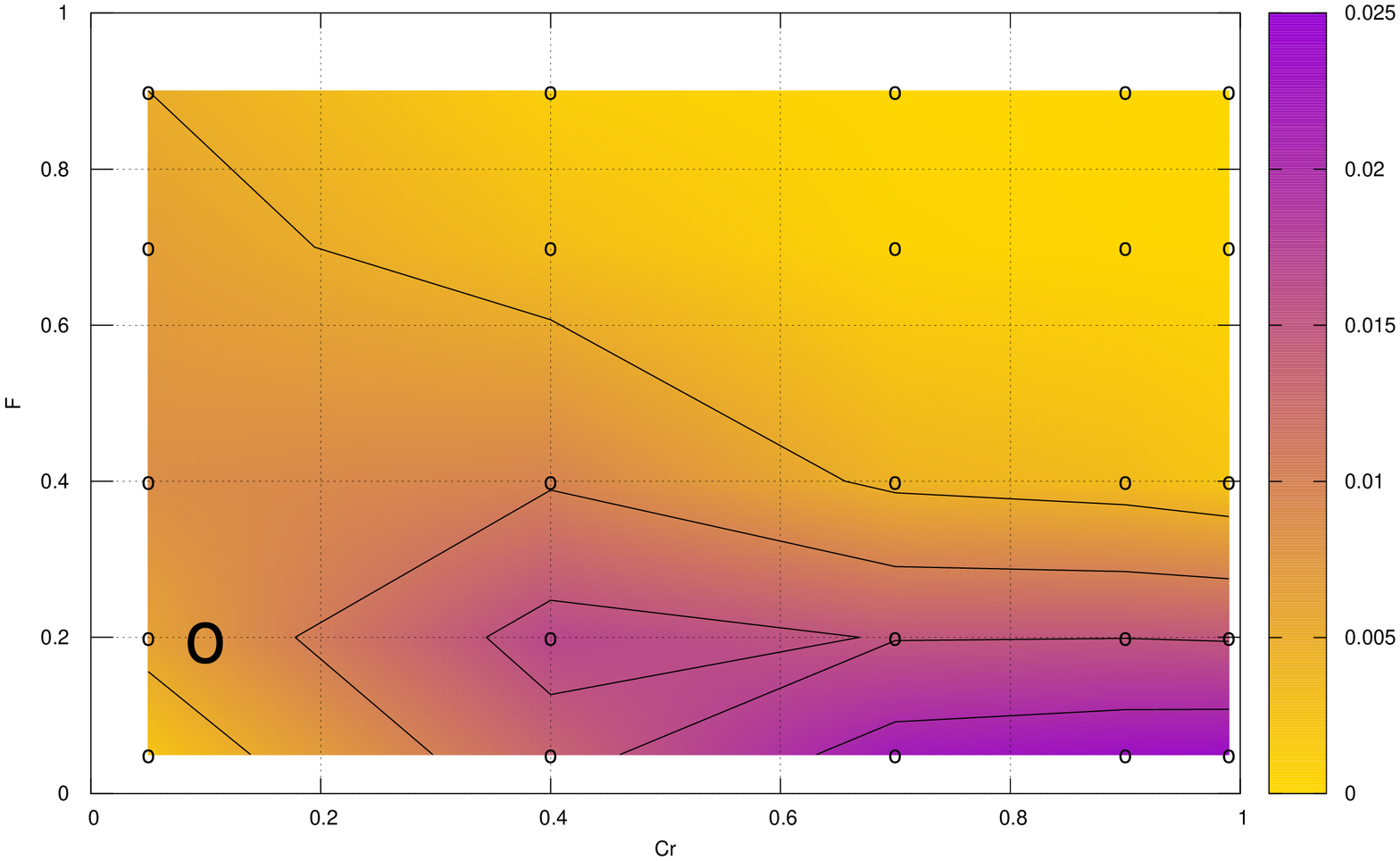}}
\caption{Example of interpolated surface of standard deviation of the percentage of corrections for {DE/rand/1/bin} for various values of $F$ and $CR$. For explanation about axes, see Section \ref{sect:teaser}.}\label{fig:stdDEROB}
\end{figure}

Similarly, Figure \ref{fig:stdDEROB} shows standard deviation of the percentage of corrections if DE/rand/1/bin configuration, over $50$ runs, for three population sizes, occurring for each one of the $25$ $F-Cr$ combination. More specifically, in each subfigure, axes pointing upwards shows values of $F$, mean while axis pointing to the right shows values of $CR$. Colour encodes the value of the a posteriori computed standard deviation of percentage of corrections for experiments with given $F$ and $CR$ values -- experimentally obtained values of standard deviation are marked with small dots, meanwhile standard deviations for the rest of $F-Cr$ couples are interpolated and smoothed. Black curves are interpolated contour curves for the standard deviation of the percentage of corrections. Colour axis is the same for all subfigures -- it runs from $0$ values in yellow to $0.35$ values in violet. Chosen $F-Cr$ paid is marked with a small circle. 

As mentioned previously, detailed analysis of the distributions, averages and standard deviations in the percentage of corrections for different configurations of DE will be published shortly as a separate publication. However, it is worth mentioning now that, unexpectedly, at least from the point of view of the analysis of the percentages of corrections, control parameters of DE seem to have different and, moreover, usually opposite effect in different configurations of DE. In other words, what our results show is that, from the point of view of dynamics, DE, in effect, is transformed into a different algorithm with the change of it's configuration -- \textit{dynamics inside DE drastically depends on the inner structure of the operators}. 

Apart from other conclusions, this also means that it is hardly ever sensible to talk about ''tuning control parameters for DE'' in general. For example, high values of $F$ do not necessarily always mean large steps of the algorithm as popular research suggests \cite{bib:Lampinen2000} -- if it was so, the number of corrections would consistently be high for all DE configurations, regardless the chosen value of $Cr$ as higher number of larger steps correlates with the higher numbers of corrections. Our results suggest that it is not the case -- higher number of corrections is indeed attained for higher values of $F$ but for some configurations this consistently happens for higher values of $Cr$ and for some -- for low values of $Cr$. Clearly, more research is required in this direction.


\section{Results on structural bias}\label{results}
Graphical results have been generated as explained in Section \ref{testingForBias} and grouped according to the specific DE scheme in the file \cite{bib:BiasDEResults} for visual inspection. To facilitate their analysis, the $9$ possible configurations of correction strategies and population size are positioned in the same page, thus giving a general view of each one of the $8$ DE configurations considered in this study.

\begin{figure}[H]
\centering
\subfigure[{Saturation correction}, ${NP{=}5}$.]{\includegraphics[width=0.347\textwidth,keepaspectratio,angle=270]{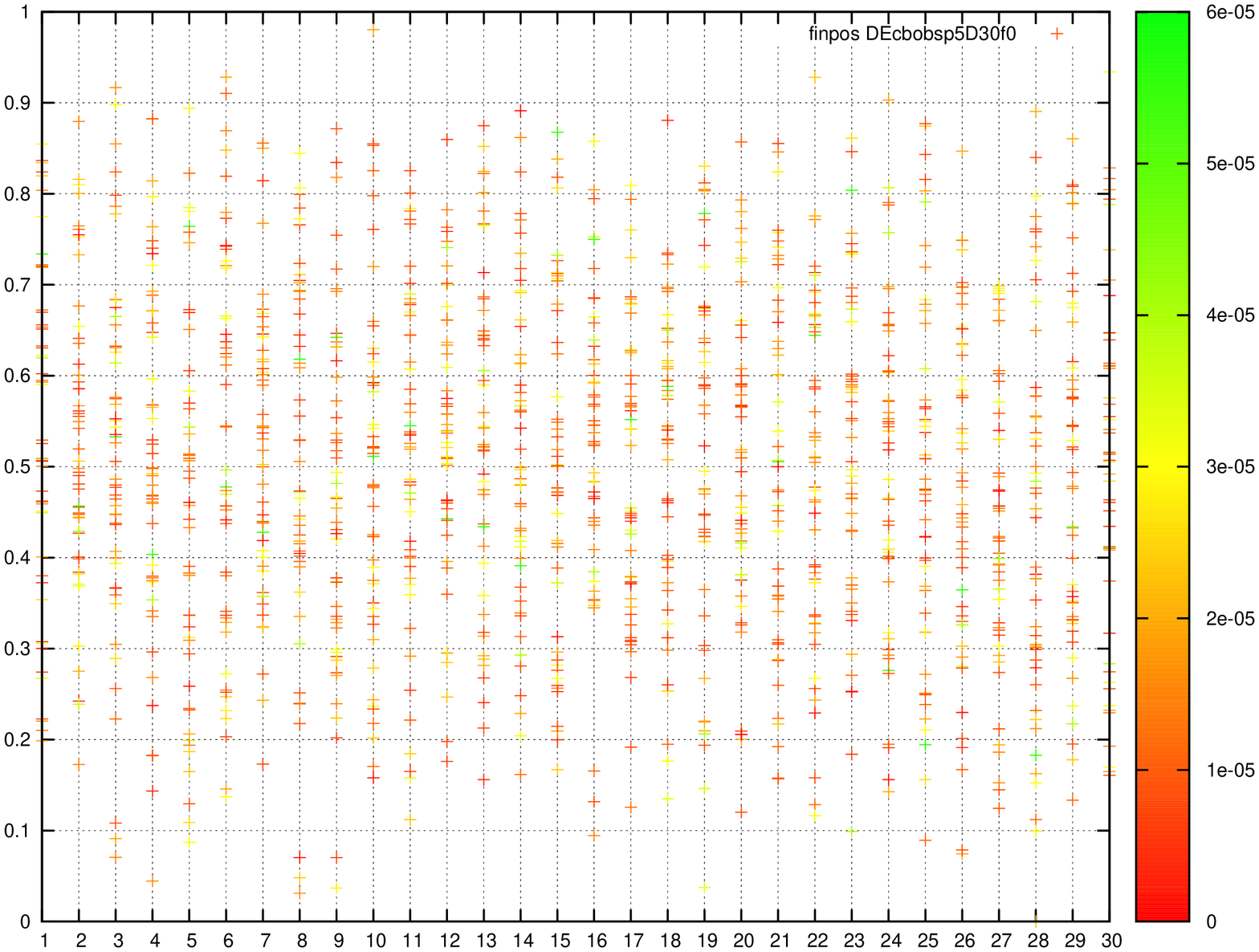}} 
\subfigure[{Toroidal correction}, ${NP{=}5}$.]{\includegraphics[width=0.347\textwidth,keepaspectratio,angle=270]{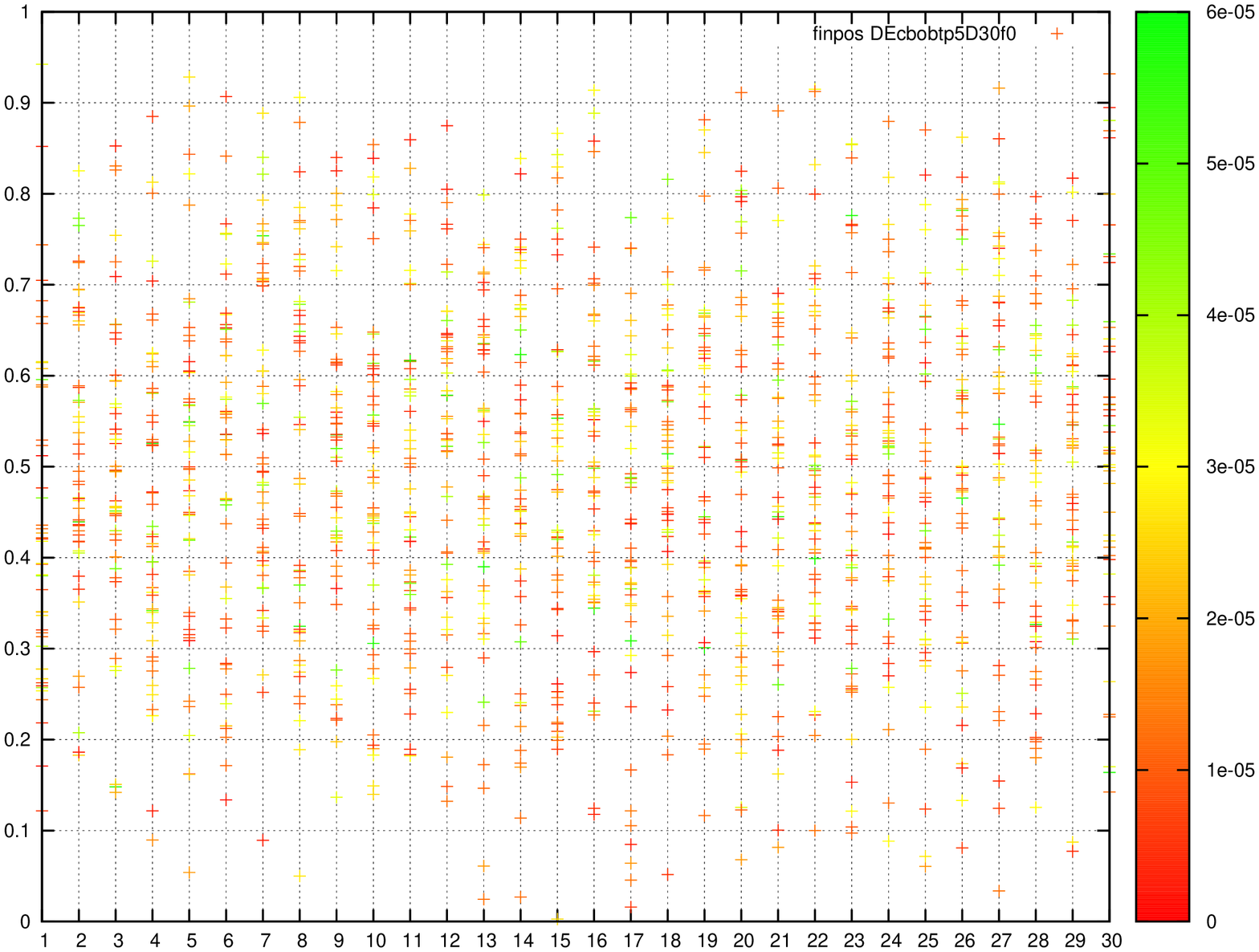}}
\subfigure[{Saturation correction}, ${NP{=}20}$.]{\includegraphics[width=0.347\textwidth,keepaspectratio,angle=270]{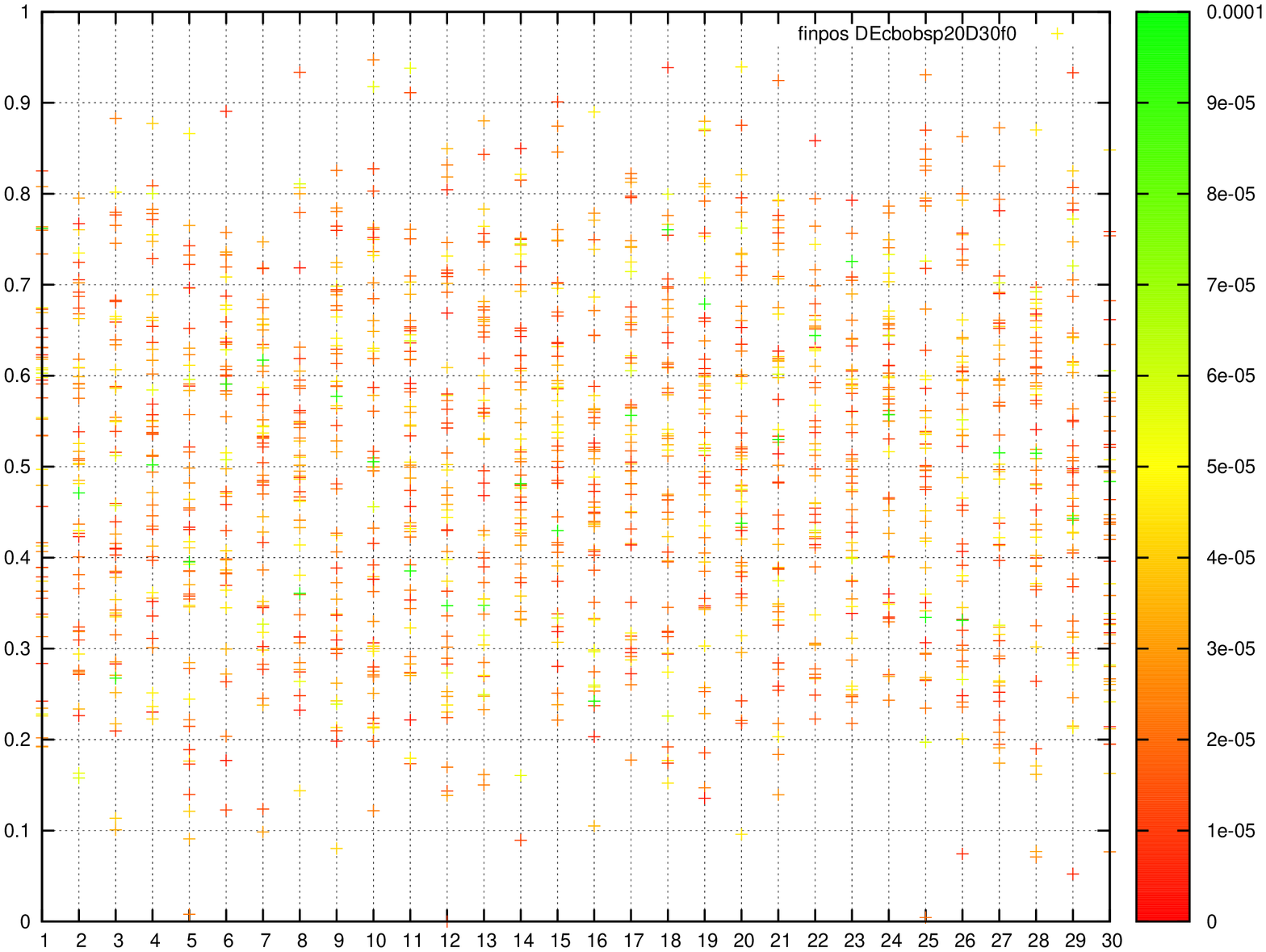}} 
\subfigure[Toroidal correction, ${NP{=}20}$.]{\includegraphics[width=0.347\textwidth,keepaspectratio,angle=270]{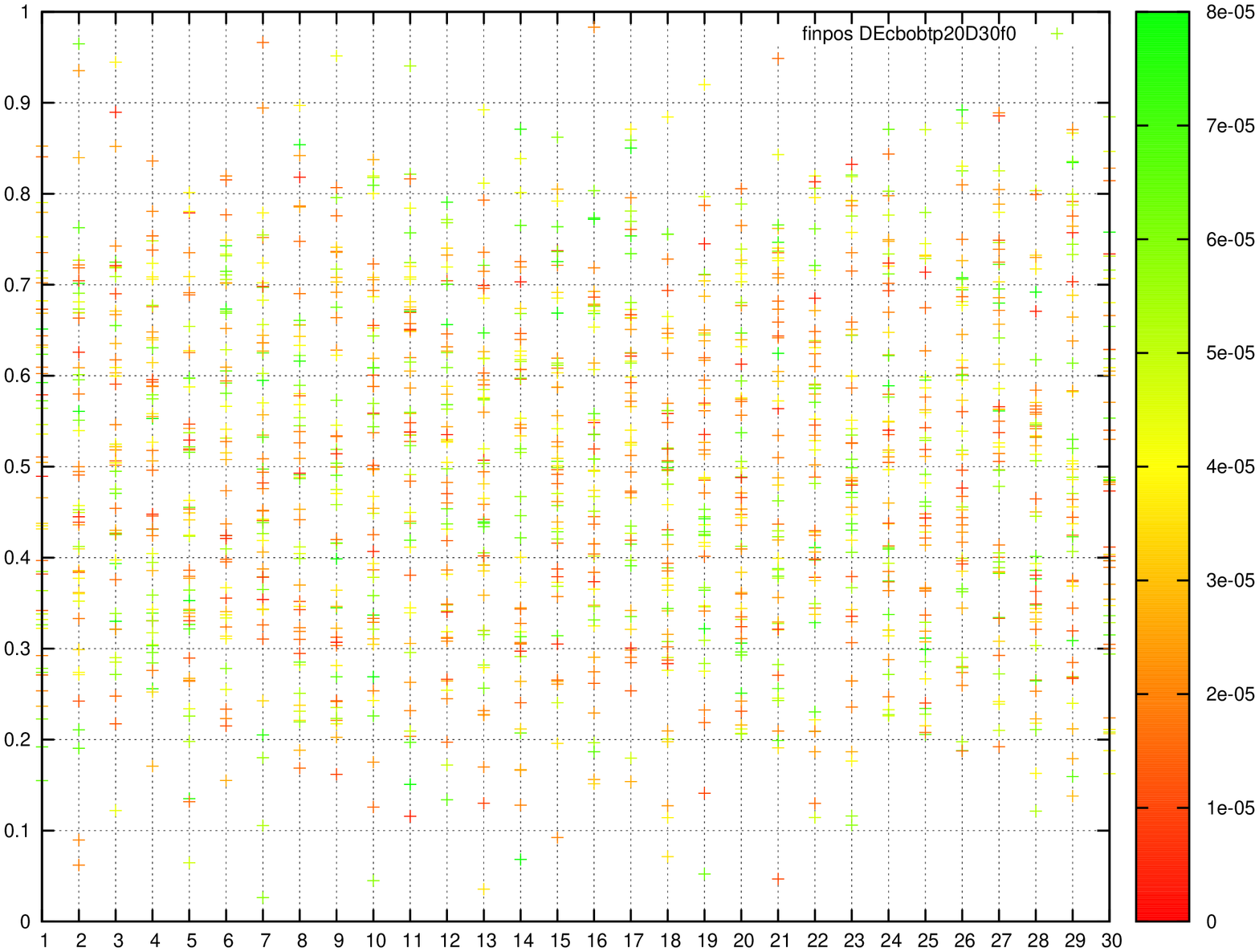}}
\subfigure[Saturation correction, ${NP{=}100}$.]{\includegraphics[width=0.347\textwidth,keepaspectratio,angle=270]{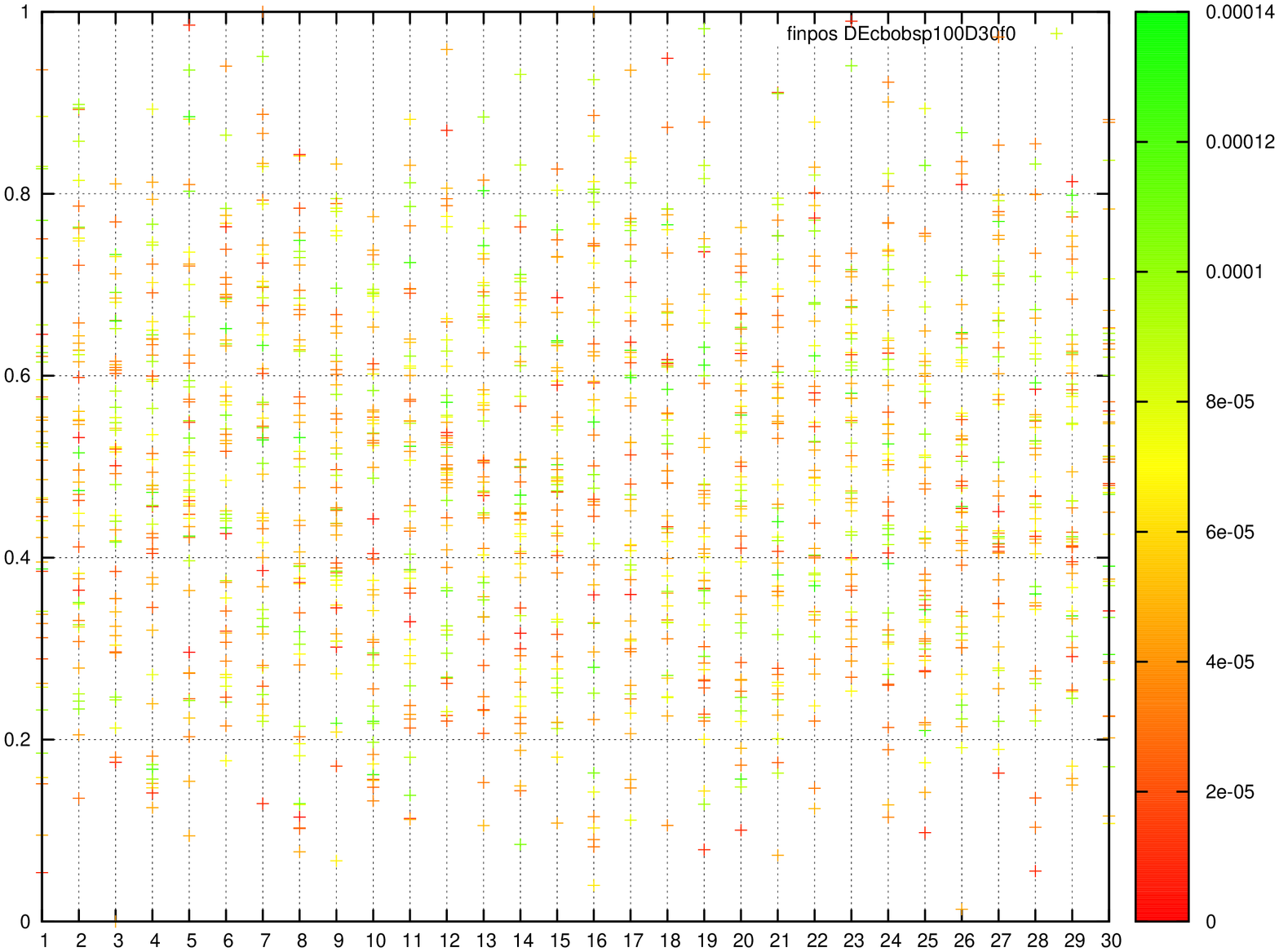}}
\subfigure[Toroidal correction, ${NP{=}100}$.]{\includegraphics[width=0.347\textwidth,keepaspectratio,angle=270]{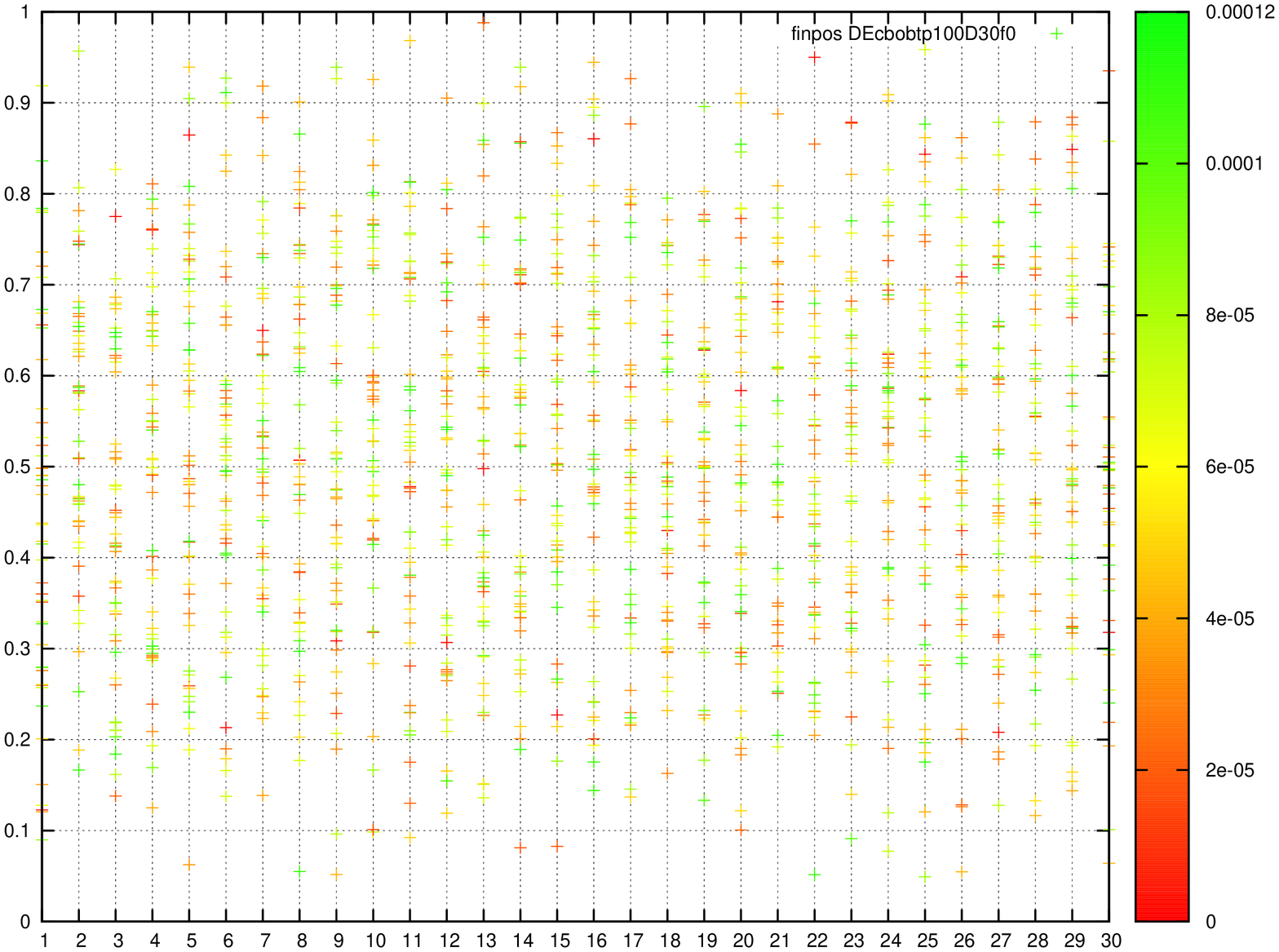}}
%
%
\caption{Results for DE/current-to-best/1/bin equipped with saturation correction (left-hand side) and with toroidal correction (right-hand side) with population size $NP$ equal to $5$ (top layer), $20$ (middle layer) $100$ (bottom layer).}\label{fig:biasedCases}
\end{figure}

\begin{figure}[H]
\centering
\subfigure[DE/rand/2/bin - saturation - , $NP=20$]{\includegraphics[width=0.347\textwidth,keepaspectratio,angle=270]{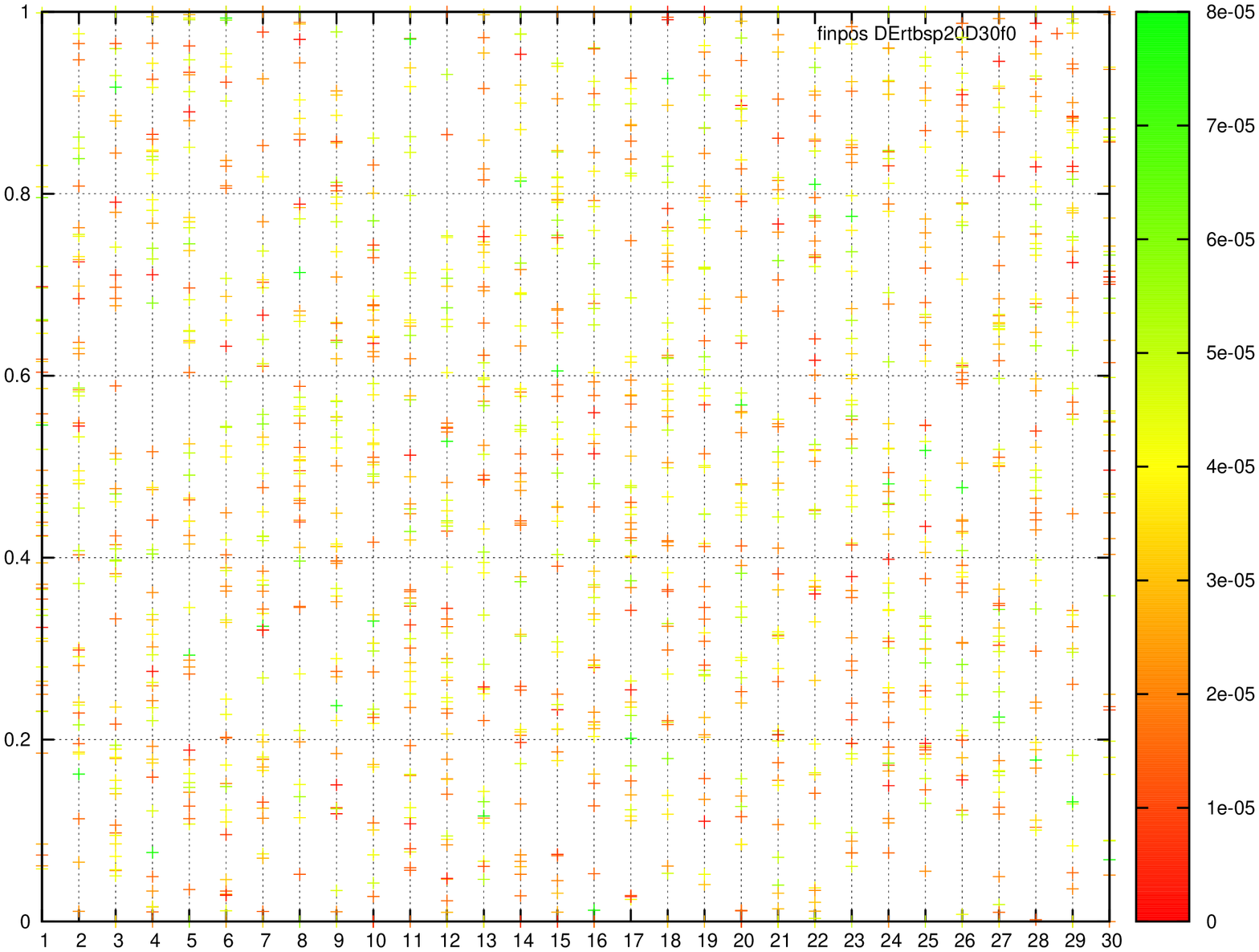}}
\subfigure[DE/rand/2/exp - saturation - , $NP=20$]{\includegraphics[width=0.347\textwidth,keepaspectratio,angle=270]{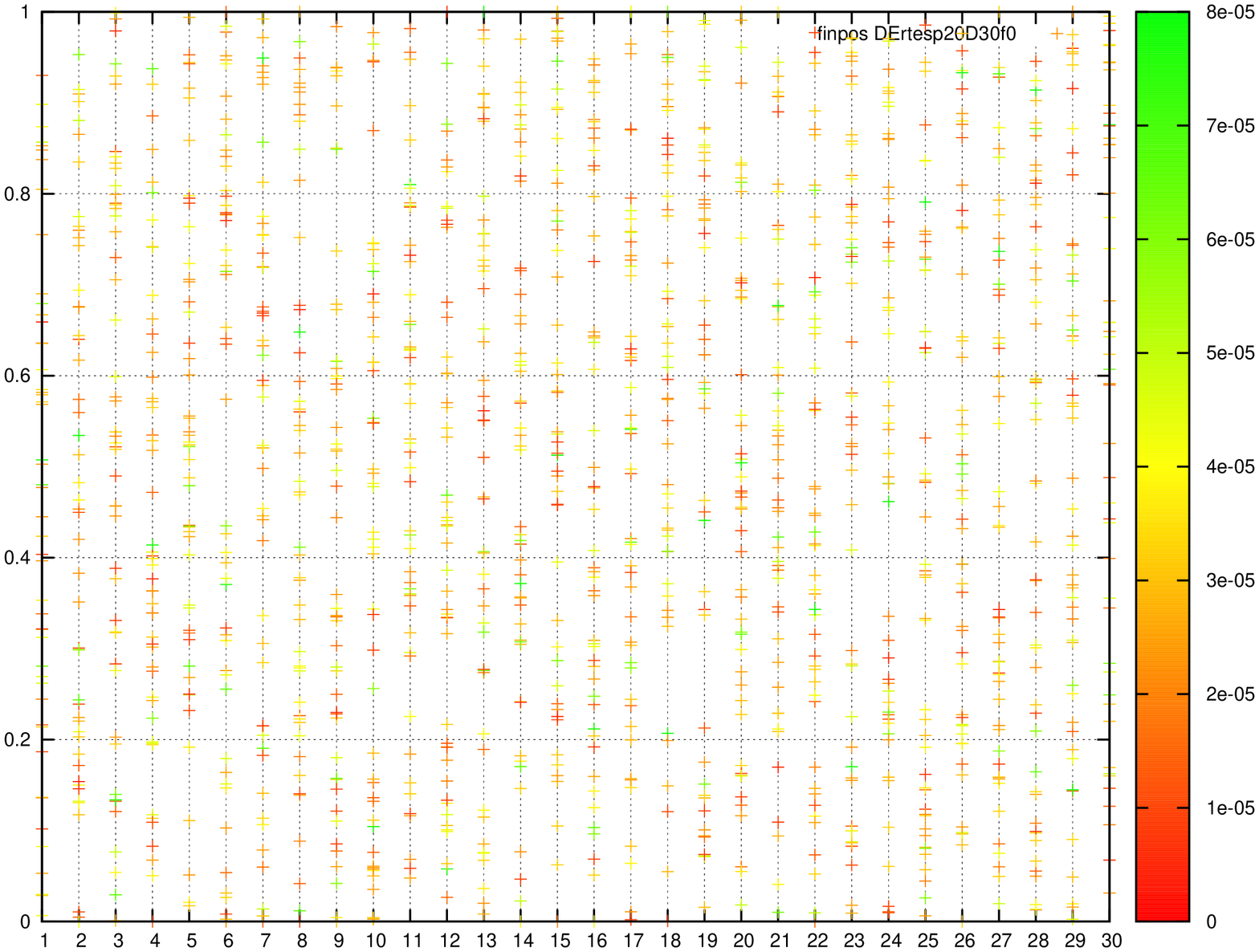}}
\subfigure[DE/current-to-best/1/exp - saturation - , $NP=5$]{\includegraphics[width=0.347\textwidth,keepaspectratio,angle=270]{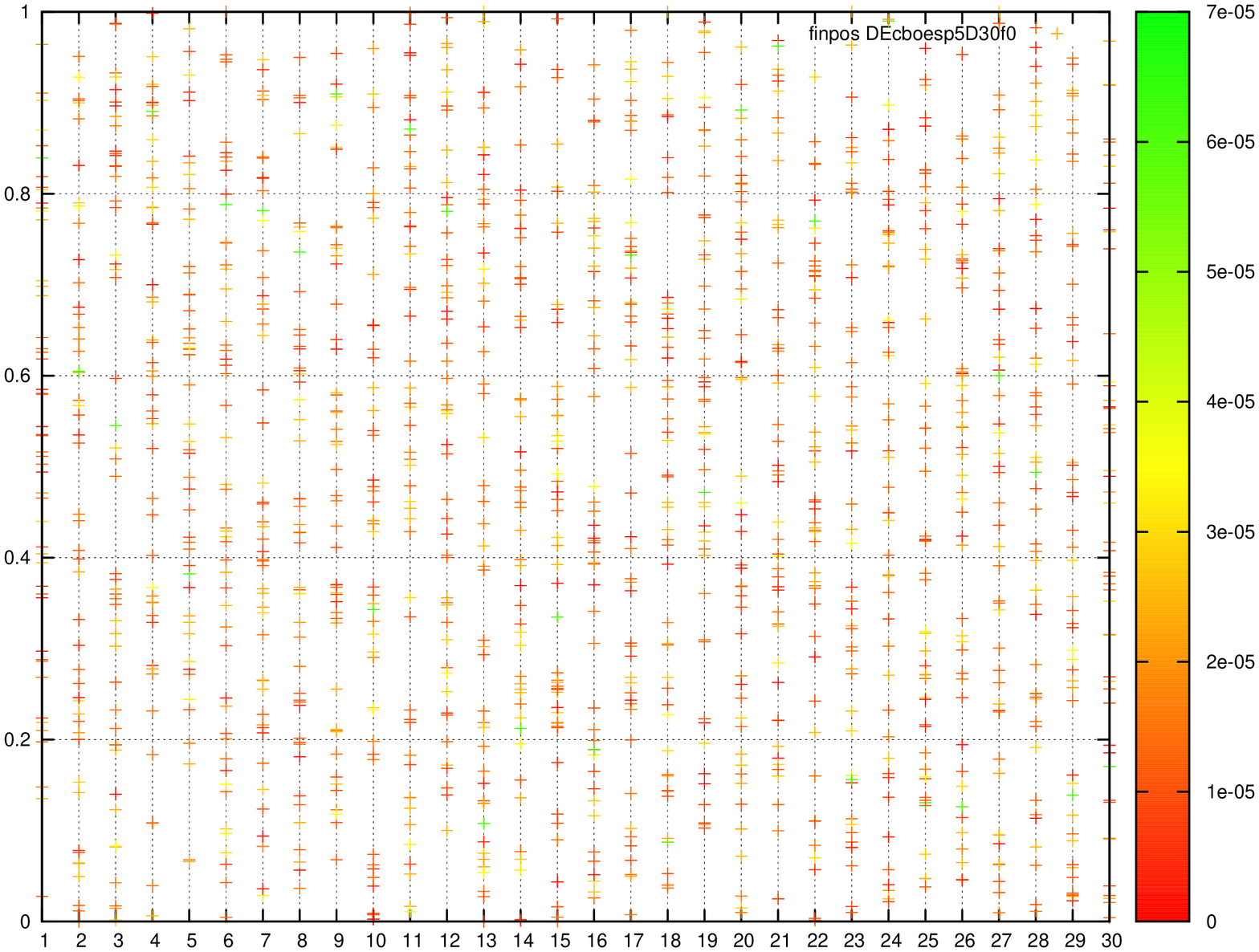}}
\subfigure[DE/current-to-best/1/exp - toroidal - , $NP=100$]{\includegraphics[width=0.347\textwidth,keepaspectratio,angle=270]{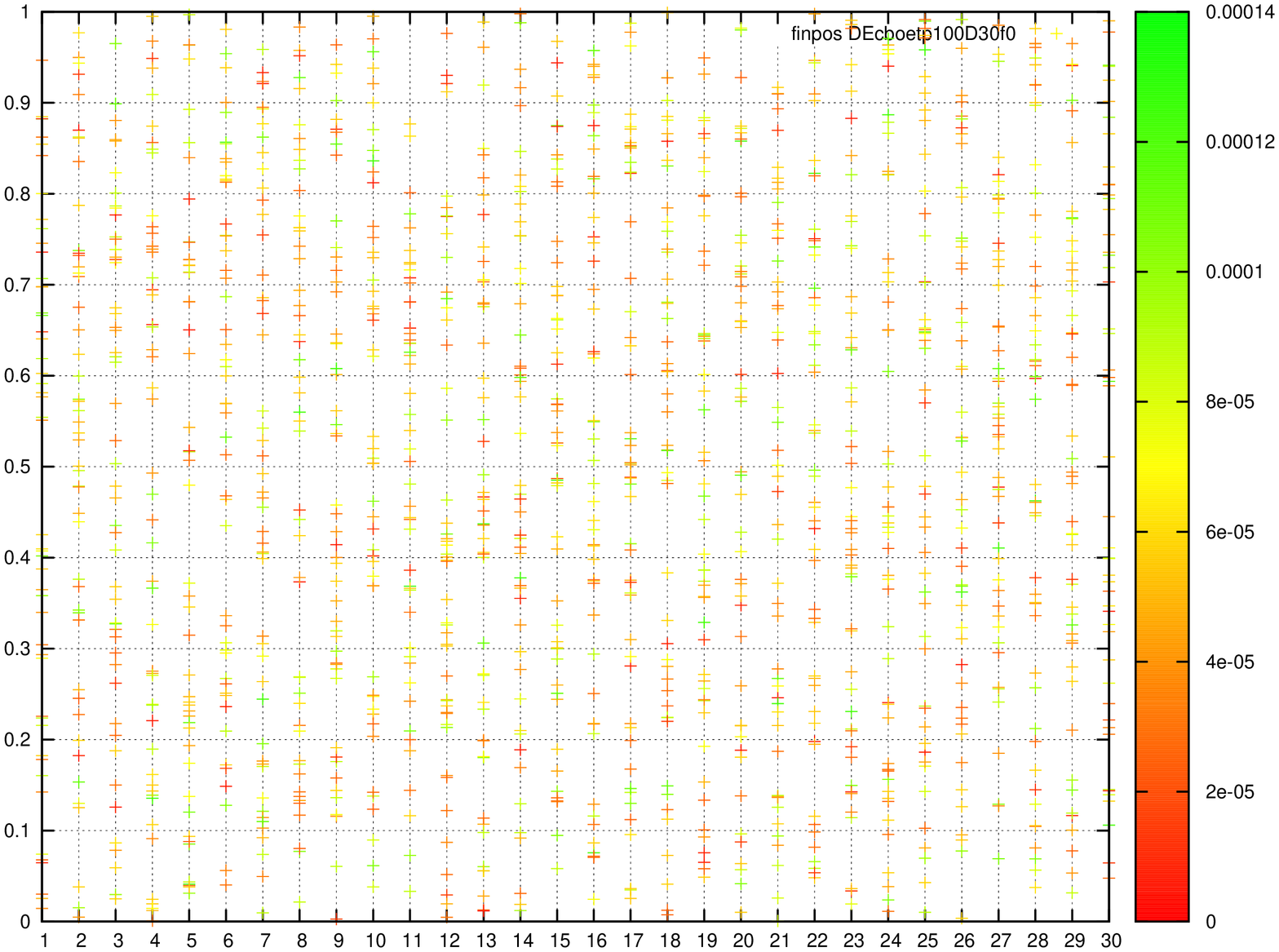}}
\caption{Results for several unbiased cases with different population sizes and different combinations of mutation, crossover, and correction strategy.}\label{fig:unbiasedCases}
\end{figure}

A glance at these results immediately brings out clear differences between DE/current-to-best/1/bin and the other schemes under examination, as it is the only configuration to display a clear structural bias. All biased patterns have been grouped Figure \ref{fig:biasedCases} in contrast to Figure \ref{fig:unbiasedCases} where some of the remaining unbiased patterns are showed.

Thus, after the first observation, one must conclude that Differential Evolution is a robust and less biased optimisation paradigm with respect to others, such as  e.g. GA and PSO examined in \cite{KONONOVA2015}. Furthermore, it appears that classic DE schemes present weaker biases than some modern self-adaptive approaches as the JADE algorithm \cite{PIOTROWSKI2018}. Most importantly, it is interesting to note that simpler mutation schemes, as DE/rand/1, DE/rand/2 and DE/best/1, do not seem to carry significant structural biases neither when the binomial crossover is employed, nor when it is replaced with the exponential one (see top row of Figure \ref{fig:unbiasedCases}). Moreover, they are not sensitive to the choice for the correction strategy and despite what observed in \cite{KONONOVA2015} -- where the strength of the structural bias increases with increasing population sizes in GA and PSO based optimisation -- the number $NP$ of individuals seems not to have a direct impact on the biases of these algorithmic structures (see bottom layer of Figure \ref{fig:unbiasedCases}).

\begin{figure}[H]
\centering
\subfigure[{Penalty correction}, ${NP{=}5}$.]{\includegraphics[width=0.347\textwidth,keepaspectratio,angle=270]{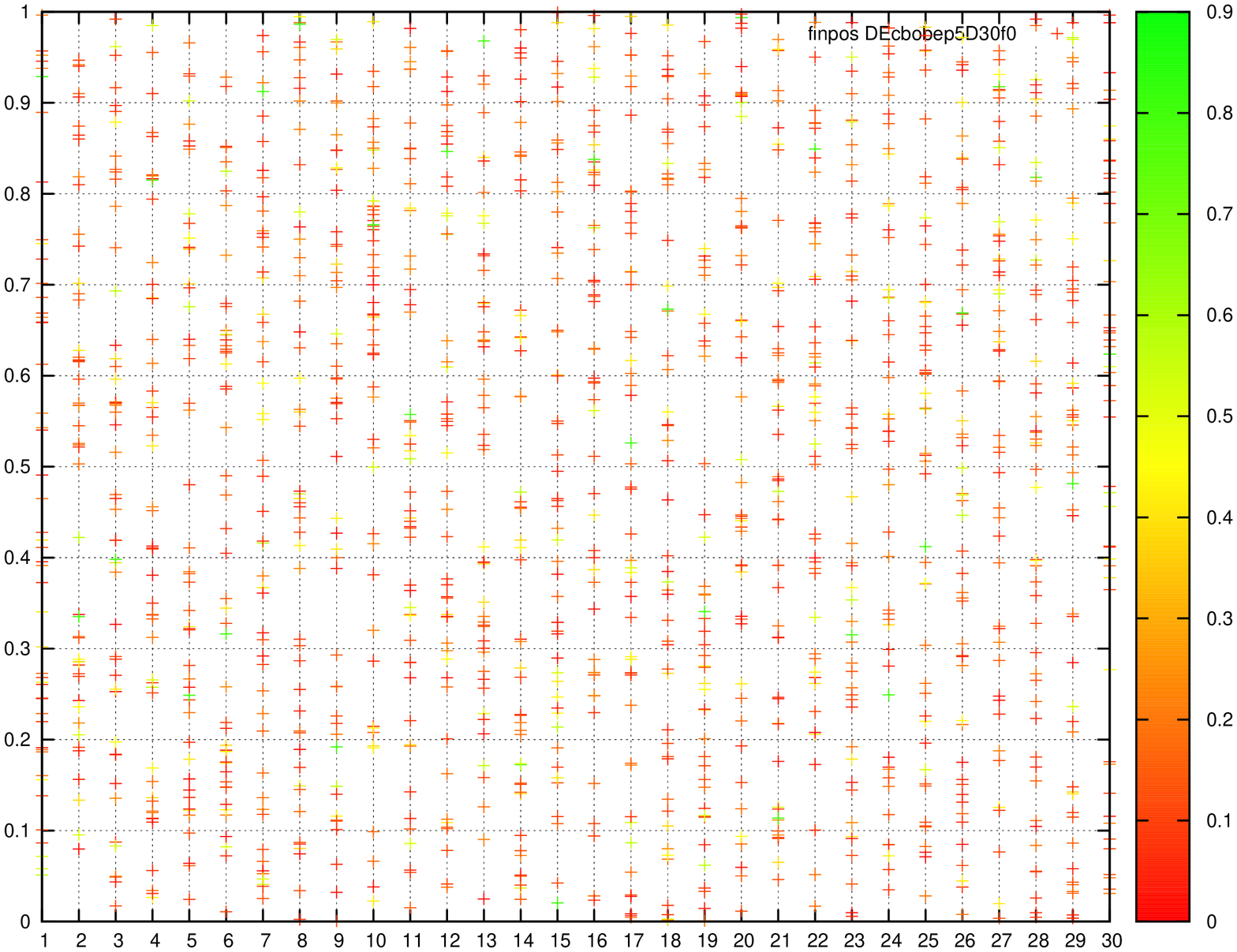}\label{sub:NP5}}
\subfigure[{penalty correction}, ${NP{=}20}$.]{\includegraphics[width=0.347\textwidth,keepaspectratio,angle=270]{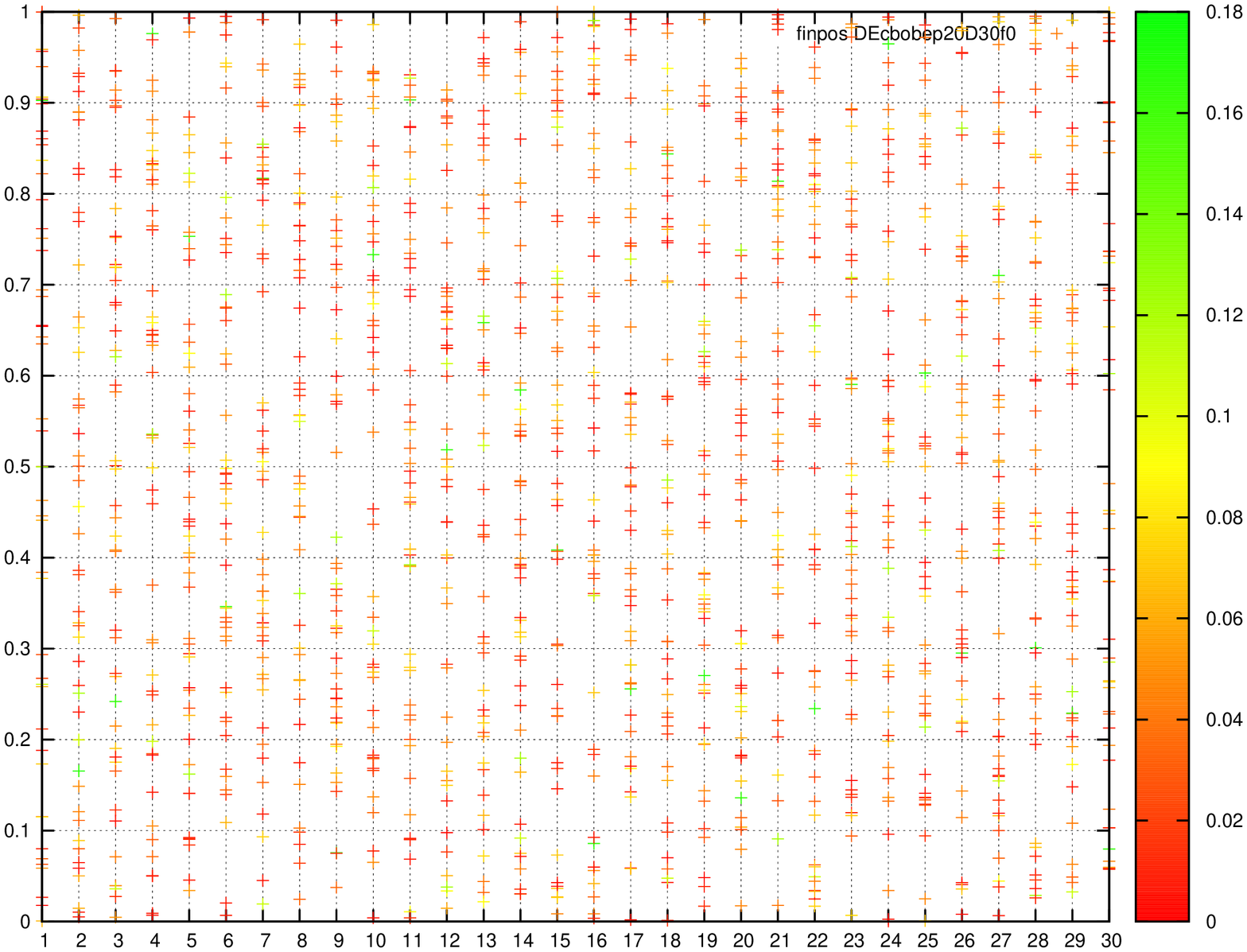}\label{sub:NP20}}
\subfigure[{Penalty correction}, ${NP{=}100}$.]{\includegraphics[width=0.347\textwidth,keepaspectratio,angle=270]{figures/finpos_DEcbobep20D30f0.eps}\label{sub:NP100}}
%
\subfigure[{GA, penalty correction}, ${NP{=}100}$.]{\includegraphics[width=0.347\textwidth,keepaspectratio,angle=270]{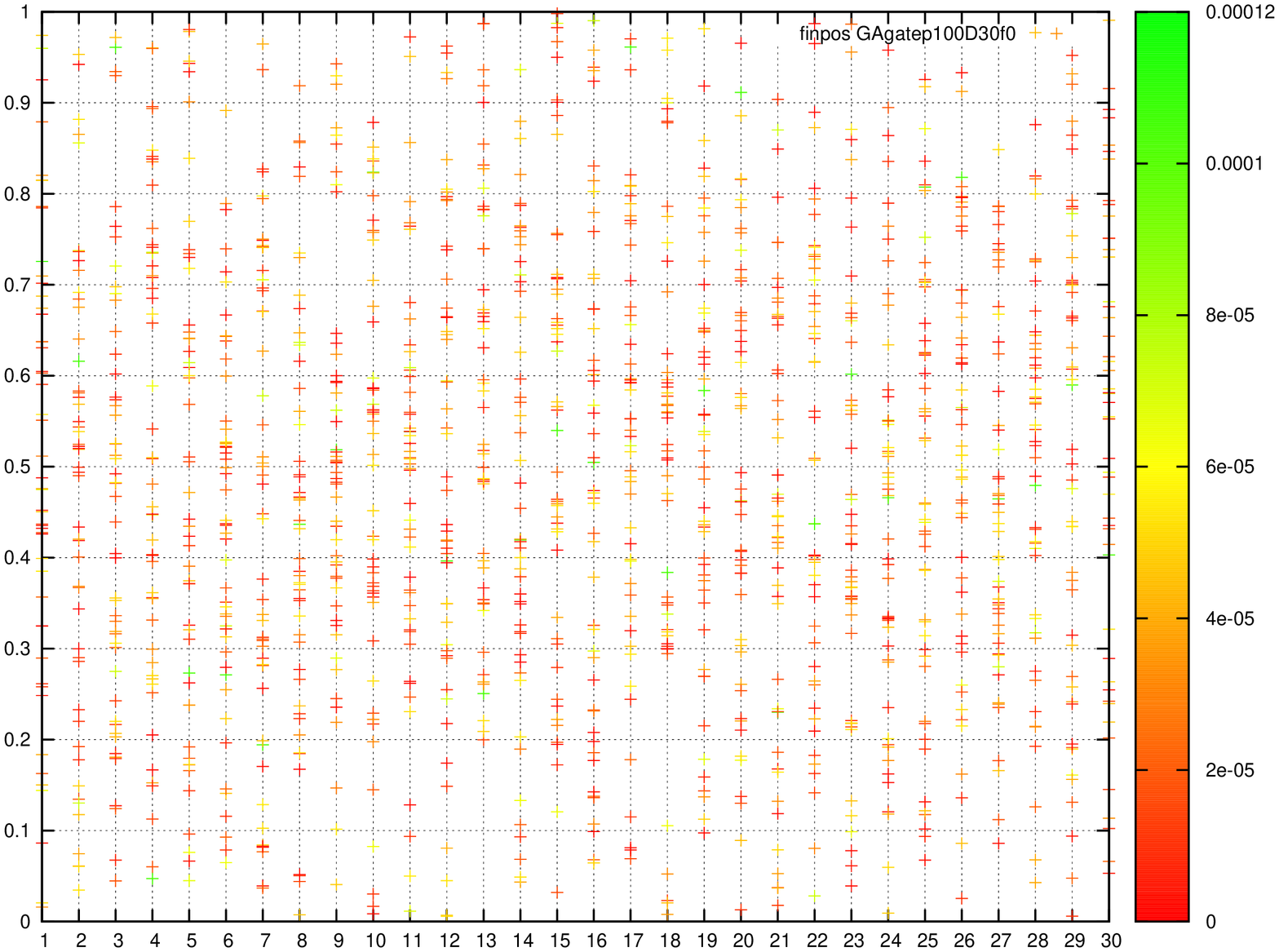}\label{sub:GA}} 
\caption{Results for DE/current-to-best/1/bin equipped with penalty correction and population size $NP=5$, see Subfigure \ref{sub:NP5}, $NP=20$, see Subfigure \ref{sub:NP20}, and $NP=100$, see Subfigure \ref{sub:NP100}. Subfigure \ref{sub:GA} shows the effect of the application of penalty correction to a Genetic Algorithm with population size $NP=100$,  Arithmetic crossover, Gaussian mutation, and tournament selection.}\label{fig:unbiasedCBOBE}
\end{figure}

Conversely, it is evident that DE/current-to-best/1/bin tends to privilege specific regions as the best individuals accumulate towards the centre of the search space at the end of the optimisation process. Once again, in contrast to what was previously observed in \cite{KONONOVA2015}, the displayed bias seems to only marginally depend on the population size, as can be seen from Figure \ref{fig:biasedCases}. It must be pointed out that even though this knowledge was already available from the preliminary study in \cite{bib:biasDELego18}, in which saturation and toroidal corrections were employed, a peculiar behaviour arose in this examination while trying the penalty approach of Formula \ref{eq:penalty}, so eliminating its structural bias (see Figure \ref{fig:unbiasedCBOBE}). It must be remarked that a similar result cannot be easily obtained with different optimisation paradigms. This is graphically shown in Figures \ref{sub:GA} which displays a visible, albeit not strong, structural bias for a GA equipped with penalty correction.


Interestingly enough, results obtained with $3$ different population sizes and $3$ different correction strategies  further confirmed the ``mitigating'' effect of the crossover operator for the structural bias of DE/current-to-best/1 \cite{bib:biasDELego18}. Indeed, see the bottom row of Figure \ref{fig:unbiasedCases}, DE/current-to-best/1/exp unexpectedly appears not to be biased at all by visual inspection. A motivation for this discrepancy can be attributed to the fact that, for a given $CR$ value, the average number of exchanged design variables is higher if the binomial crossover is used, whereas it is lower in presence of the exponential crossover. The latter logic is indeed based on a different working principle according to which the probability of exchanged components decays exponentially and depends on the dimension $n$ of the problem -- a fair comparison between two DE algorithms employing bin and exp crossovers should consider the ad-hoc crossover rates $CR_{bin}=CR$ and $CR_{exp}=\frac{1}{\sqrt[nCR]{2}}$ respectively \cite{rot2018}. This means that, for the considered dimension value $n=30$, only a few components are in fact exchanged. One can therefore speculate that such a low value is simply not enough to manifest a visible structural bias. However, this should not be the case as it would implies that only the crossover operator, and in particular the binomial one, introduces biases into the DE framework while all considered mutation strategies are unbiased. At the current state of this investigation, this does not seem to be reasonable considering that both the bin and exp options have been used with other mutation schemes in this study but clear biases, visible by the naked eye, were not individuated. 

The latter considerations suggest that the structural bias of an optimisation algorithm is not  simply the result of the sum of biases of its operators and can also arise under particular mutation-crossover combinations, or in general, under specific combination of both biased and unbiased operators. In this light, the structural bias can also be seen as a non-linear phenomenon where the biases of two biased operators does not necessarily add up if they are combined within an algorithm, but rather generate spurious contributions biasing the search further and differently.

\section{Conclusions}\label{conclusions}
This paper systematically analyses a wide range of popular DE configurations in the light of structural bias that manifest itself during optimisation process as the algorithm ``preferring'' some regions of the search space to the others. Such behaviours has been clearly demonstrated in other EA algorithms and shown to stem from iterative application of algorithmic operators, each potentially plagued with their own shortcomings. These shortcomings are thus superimposed by the algorithm's structure and could thus limit the overall performance of the algorithm due to an uneven exploration of the search space. 

Compared to previous studies, a new aspect usually overlooked during algorithmic design stage is considered as potential contributor to the formation of structural bias -- choice of constraint handling technique --  which has eventually turned out to be crucial in mitigating the effects of structural bias on an otherwise highly biased DE configuration. This is evident in DE/current-to-best/1/bin whose bias can be eliminated by simply employing the penalty constraint handling strategy. In this light, practitioners are recommended to use DE/current-to-best/1/bin in combinations with such constraint handling method, and also algorithm designers to pay more attention to the often underestimated details which could, surprisingly, make the big difference.

Triggering mechanisms of structural bias in DE has also turned out to be different from that of GA and PSO -- population size has no clear effect on the strength of structural bias. Main results suggest that the structural bias of an optimisation algorithm is really a superposition and not a sum of effects of individual operators as in general it can arise from a specific combination of both biased and unbiased operators generating spurious contributions and biasing the search further and in a different manner. 

In the future, promising areas of investigations are then two-folded: first, to examine the contribute of algorithmic operators (in support of our observation that the system as a whole can be biased regardless of the single component), their structural bias has to be searched and quantified; second, alternative ways to visualise/quantify bias, e.g. clustering, can be designed to flag critical cases and confirm the absence/presence of structural bias. These would clarify unsolved matters and helps understand the triggering  mechanism for the bias, not only in DE. 

Results presented in this paper have also suggested an interesting direction for future research -- interpretation of dynamics inside the DE population based on proxy estimates and subsequent interpretation of the control parameters pair. To the best of our knowledge, to date, such interpretation valid for all DE configurations is still missing.

\bibliographystyle{elsarticle-harv} 
\bibliography{refs}
\end{document}

%% file: MISC/newCommandsAndPackages.tex
\usepackage[dvips]{epsfig,graphicx}
\usepackage{times,subfigure,longtable,lscape,amssymb,amsbsy,times,fancyhdr,color,amsmath,latexsym,rotating,hhline,url,multirow} 
\usepackage{float,amsmath,graphicx,latexsym,amsfonts,amssymb,soul,wrapfig,dsfont}
\usepackage{caption,algorithm,algpseudocode}
\usepackage{textcomp}
\usepackage{hyperref}

\usepackage{float}
\usepackage{footnote}
\makesavenoteenv{tabular}
\makesavenoteenv{table}

\usepackage{pifont}
\newcommand{\cmark}{\ding{51}}%
\newcommand{\xmark}{\ding{55}}%


\setcounter{MaxMatrixCols}{10}

\algdef{SE}[DOWHILE]{Do}{doWhile}{\algorithmicdo}[1]{\algorithmicwhile\ #1}%

 \usepackage{tikz,pgfplots}
\pgfplotsset{compat=1.14}
\usepackage{gnuplottex}

%% file: TABS/corrStrat.tex
%
%
%
%
\begin{table}[ht!]
\small
\caption{Advantages and disadvantages of common constraints handling techniques used to deal with solutions generated outside the domain} \label{tab:corr_strategies}
\centering
\begin{tabular}{c|p{95mm}|r}	
\hline \hline
\multirow{3}{*}{\textbf{Dismiss}}  & \textbullet\hspace{0.05cm}  Unaltered objective function & \cmark \\ 
\cline{2-3}
 &\textbullet\hspace{0.05cm}  Wasteful of computational resources & \multirow{2}{*}{\xmark}\\
 & \textbullet\hspace{0.05cm}  Can easily lead to no result &\\
\hline \hline

\multirow{6}{*}{\textbf{Penalise}}  &
 \textbullet\hspace{0.05cm}  Traditional and ''cleanest'' way to deal with constraints & \cmark\\ 
 \cline{2-3}
 &\textbullet\hspace{0.05cm} Allows unfeasible individuals (problematic in applications) & \multirow{4}{*}{\xmark}\\
&  \textbullet\hspace{0.05cm} Can result in an infeasible result &\\
&  \textbullet\hspace{0.05cm}  Can be viewed as a distortion of the original objective function &\\
&  \textbullet\hspace{0.05cm} Introduces extra design choices requiring investigation and justification\footnote{Choice of penalty function depends on (1) the ration between sizes of the feasible and the whole search space, (2) the topological properties of the feasible search space, (3) the type of objective function, (4) the number of variables, (5) the number of constraints, (6) types of constraints, (7) number of active constraints at the optimum \citep{bib:Michalewicz1995}.} &\\
&  \textbullet\hspace{0.05cm} Requires knowing the range of values of the objective function  &\\
\hline \hline
\multirow{4}{*}{\textbf{Correct}}  &  \textbullet\hspace{0.05cm} Very straightforward & \multirow{2}{*}{\cmark}\\  &\textbullet\hspace{0.05cm}  Provides additional moves to the algorithm facilitating diversity &\\ 
\cline{2-3}
 &\textbullet\hspace{0.05cm} Distorts objective function if the correction operator is biased &\multirow{2}{*}{\xmark} \\

 &\textbullet\hspace{0.05cm}  Introduces extra decisions that need to be taken by algorithm designer, e.g.: which particular correction method to choose; whether or not to inherit the corrected genotype; in case of using the local search, whether or not to use the greedy algorithm and in what order to examine the variables. &\\
\hline \hline
\end{tabular}
\end{table}

%% file: PSEUDOCODES/DE.tex
\begin{algorithm}[H]
 \caption{Differential Evolution}\label{alg:DE}
\begin{center}
\begin{algorithmic}[0]
\State $g\gets 1$\Comment{First generation}
\State $\mathbf{Pop^g}\gets$ \textit{randomly sample $NP$ individuals within the search space \textbf{ D }$\subset\mathbb{R}^n$}
\State $\mathbf{x_{best}}\gets$\textit{fittest individual}$\in \mathbf{Pop^g}$
\While{\textit{condition on budget}}
	\For{\textit{each} $\mathbf{x}\in\,\mathbf{Pop^g}$}
		\State $\mathbf{x_m}\gets$\textit{Mutation} \Comment{e.g. Formula \ref{eq:derand1} in \cite{bib:Storn1995}} 
		\State $\mathbf{x_{offspring}}\gets$\textit{CrossOver}$\left(\mathbf{x},\mathbf{x_m}\right)$\Comment{e.g. Algorithm \ref{alg:xobin} in \cite{bib:Storn1995}}
		\If{$f\left(\mathbf{x_{offspring}}\right)\leq f\left(\mathbf{x}\right)$} 
			\State $\mathbf{Pop^{g+1}}\gets\mathbf{x_{offspring}}$
		\Else\Comment{Fill the new population for the next iteration}
			\State $\mathbf{Pop^{g+1}}\gets \mathbf{x}$
		\EndIf
	\EndFor
	\State $g\gets g+1$\Comment{Replace the old with the new generation}
	\State $\mathbf{x_{best}}\gets$ \textit{fittest individual}  $\in \mathbf{Pop^g}$\Comment{update best individual}
\EndWhile
\State \textbf{Output} \textit{Best Individual} $\mathbf{x_{best}}$
\end{algorithmic}
\end{center}
\end{algorithm}

%% file: MISC/DEMutationOperators.tex
\begin{itemize}
 \item {rand/1: \begin{equation}\label{eq:derand1}
           					    	\mathbf{x_m} = \mathbf{x_{r_1}}+F\left(\mathbf{x_{r_2}} - \mathbf{x_{r_3}}\right)
        		  \end{equation}}
    \item { rand/2: \begin{equation}\label{eq:derand2}
         					 	\mathbf{x_m} = \mathbf{x_{r_1}}+F\left(\mathbf{x_{r_2}} - \mathbf{x_{r_3}}\right) + F\left(\mathbf{x_{r_4}}-\mathbf{x_{r_5}}\right)
        			  \end{equation}}
    \item {best/1: \begin{equation}\label{eq:debest1}
            					\mathbf{x_m} = \mathbf{x_{best}}+F\left(\mathbf{x_{r_1}} - \mathbf{x_{r_2}}\right)
        		     \end{equation}}
    \item {current-to-best/1:\begin{equation}\label{eq:decurtobest}
            		                \mathbf{x_m} = \mathbf{x}+F \left(\mathbf{x_{best}} - \mathbf{x}\right)+F \left(\mathbf{x_{r_1}}-\mathbf{x_{r_2}}\right)
            		           \end{equation}}
\end{itemize}

%% file: PSEUDOCODES/bin.tex
\begin{algorithm}[H]
\caption{Binomial crossover}\label{alg:xobin}
\begin{center}
\begin{algorithmic}[0]
\State \textbf{Input} two parents $\mathbf{x_1}$ and $\mathbf{x_2}$\Comment{$\mathbf{x_1},\mathbf{x_2}\in$\textbf{ D }$\subset\mathbb{R}^n$}
\State $\mathbf{x_{offspring}}\gets\mathbf{x_1}$
\State $\text{Index}\gets \mathcal{I}$\Comment{$\mathcal{I}$ is uniformly sampled in $\left[1,n\right]\subset\mathbb{N}$}
\For{$i=1,\dots,n$}
     \If{$\mathcal{U}\leq CR$ or $i=\text{Index}$} \Comment{$\mathcal{U}$ is uniformly sampled in $\left[0,1\right]\subset\mathbb{R}$}
		\State $\mathbf{x}^{(i)}_{\mathbf{offspring}}\gets\mathbf{x}^{(i)}_\mathbf{2}$\Comment{exchange the $i^{th}$ component}
	\EndIf
\EndFor
\State \textbf{Output} $\mathbf{x_{offspring}}$
\end{algorithmic}
 \end{center}
\end{algorithm}

%% file: PSEUDOCODES/exp.tex
\begin{algorithm}[H]
   \caption{Exponential crossover}\label{alg:xoexp}
\begin{center}
\begin{algorithmic}[0]
\State \textbf{Input} two parents $\mathbf{x_1}$ and $\mathbf{x_2}$\Comment{$\mathbf{x_1},\mathbf{x_2}\in$\textbf{ D }$\subset\mathbb{R}^n$}
\State $\mathbf{x_{offspring}}\gets\mathbf{x_1}$
\State $i,\text{Index}\gets \mathcal{I}$\Comment{$\mathcal{I}$ is uniformly sampled in $\left[1,n\right]\subset\mathbb{N}$}
\Do
\State $\mathbf{x}^{(i)}_\mathbf{offspring}\gets\mathbf{x}^{(i)}_\mathbf{2}$\Comment{exchange the $i^{th}$ component}
\State $i\gets i+1$
\If{$i>n$}
			\State $i\gets 1$
		\EndIf
\doWhile{$\mathcal{U}\leq CR$ and $i\neq \text{Index}$} \Comment{$\mathcal{U}$ is uniformly distributed in $\left[0,1\right]\subset\mathbb{R}$}
\State \textbf{Output} $\mathbf{x_{offspring}}$
\end{algorithmic}
\end{center}
\end{algorithm}

%% file: PSEUDOCODES/saturation.tex
\begin{algorithm}[H]
\caption{Saturation correction}\label{alg:saturation}
\begin{center}
\begin{algorithmic}[0]
\State \textbf{Input} a solution $\mathbf{x}$ and problem's boundaries \Comment{$\mathbf{x}\in\mathbb{R}^n$ but domain is $\textbf{ D }\subset\mathbb{R}^n$}
\For{$i=1,\dots,n$}
	\State {l}$\gets${$i^{th}$ lower bound}
	\State {u}$\gets${$i^{th}$ upper bound}
	\If{$\mathbf{x}^{(i)} >$\textit{u}}
		\State $\mathbf{x}^{(i)}_\mathbf{saturated}\gets$ u\Comment{saturate the $i^{th}$ component to the upper-bound}
	\ElsIf{$\mathbf{x ^{(i)}}<$\textit{l}}
		\State $\mathbf{x}^{(i)}_\mathbf{saturated}\gets$ l \Comment{saturate the $i^{th}$ component to the lower-bound}
	\Else
		\State $\mathbf{x}^{(i)}_\mathbf{saturated}\gets\mathbf{x}^{(i)}$\Comment{keep the original $i^{th}$ component}
	\EndIf		
\EndFor
\State \textbf{Output} $\mathbf{x_{saturated}}$
\end{algorithmic}
\end{center}
\end{algorithm}

%% file: PSEUDOCODES/toroidal.tex
\begin{algorithm}[H]
\caption{Toroidal correction}\label{alg:toro}
\begin{center}
\begin{algorithmic}[0]
\State \textbf{Input} a solution $\mathbf{x}$ and problem's boundaries \Comment{$\mathbf{x}\in\mathbb{R}^n$ but domain is $\textbf{ D }\subset\mathbb{R}^n$}
\For{$i=1,\dots,n$}
	\State {l}$\gets${$i^{th}$ lower bound}
	\State {u}$\gets${$i^{th}$ upper bound}
	\State $x_N\gets\frac{\mathbf{x}^{(i)}-\textit{l}}{u-l}$\Comment{normalise $i^{th}$ component of $\mathbf{x}$ in $[0,1]$}
	\State $x_{R}\gets${rounds} $\mathbf{x}^{(i)}$ {to the nearest integer towards $0$}
	\If{$x_N > 1$}
		\State $x_N\gets x_N-x_R$
	\ElsIf{$x_N < 0$}
		\State $x_N\gets 1 - \left|x_N-x_R\right|$
	\EndIf	
	\State $\mathbf{x_{corrected}}^{(i)}\gets{l}+X_N\cdot\left(u-l\right)$\Comment{scale the corrected value back to \textbf{D}}
\EndFor
\State \textbf{Output} $\mathbf{x_{corrected}}$
\end{algorithmic}
\end{center}
\end{algorithm}

%% file: PSEUDOCODES/discard.tex
\begin{algorithm}[H]
\caption{Discard correction}\label{alg:discard}
\begin{center}
\begin{algorithmic}[0]
\State \textbf{Input} a solution $\mathbf{x}$ and problem's boundaries \Comment{$\mathbf{x}\in\mathbb{R}^n$ but domain is $\textbf{ D }\subset\mathbb{R}^n$}
\If{$\mathbf{x}\not\in\mathbf{D}$}
    \State $\mathbf{x}\gets$ a selected parent
\EndIf
\State \textbf{Output} $\mathbf{x}$ \Comment{No fitness functional call required}
\end{algorithmic}
\end{center}
\end{algorithm}